\documentclass[sigconf]{acmart}

\renewcommand\footnotetextcopyrightpermission[1]{} 
\makeatletter
\renewcommand\@formatdoi[1]{\ignorespaces}
\makeatother

\usepackage{multirow}
\usepackage{subfigure} 
\AtBeginDocument{%
  \providecommand\BibTeX{{%
    \normalfont B\kern-0.5em{\scshape i\kern-0.25em b}\kern-0.8em\TeX}}}

\acmSubmissionID{926}

\begin{document}

\title{SM-SGE: A Self-Supervised Multi-Scale Skeleton Graph Encoding Framework for Person Re-Identification}



\author{Haocong Rao$^{1,2}$, Xiping Hu$^{1,2,3}$\footnotemark[1], Jun Cheng$^{1,2}$, Bin Hu$^{4,3}$\authornotemark}
\affiliation{%
  \institution{$^1$Shenzhen Institute of Advanced Technology, Chinese Academy of Sciences}
    \institution{$^2$The Chinese University of Hong Kong, Hong Kong}
      \institution{$^3$Lanzhou University}
        \country{$^4$Beijing Institute of Technology}
}
\email{haocongrao@gmail.com, huxp@lzu.edu.cn, jun.cheng@siat.ac.cn,  bh@bit.edu.cn}
\authornote{Corresponding authors}







\renewcommand{\shortauthors}{Rao and Hu, et al.}

\begin{abstract}
Person re-identification via 3D skeletons is an emerging topic with great potential in security-critical applications. Existing methods typically learn body and motion features from the body-joint trajectory, whereas they lack a systematic way to model body structure and underlying relations of body components beyond the scale of body joints. In this paper, we for the first time propose a Self-supervised Multi-scale Skeleton Graph Encoding (SM-SGE) framework that comprehensively models human body, component relations, and skeleton dynamics from \textit{unlabeled} skeleton graphs of various scales to learn an effective skeleton representation for person Re-ID. Specifically, we first devise \textit{multi-scale skeleton graphs} with coarse-to-fine human body partitions, which enables us to model body structure and skeleton dynamics at multiple levels. Second, to mine inherent correlations between body components in skeletal motion, we propose a \textit{multi-scale graph relation network} to learn structural relations between adjacent body-component nodes and collaborative relations among nodes of different scales, so as to capture more discriminative skeleton graph features. Last, we propose a novel \textit{multi-scale skeleton reconstruction mechanism} to enable our framework to encode skeleton dynamics and high-level semantics from unlabeled skeleton graphs, which encourages learning a discriminative skeleton representation for person Re-ID. Extensive experiments show that SM-SGE outperforms most state-of-the-art skeleton-based methods. 
We further demonstrate its effectiveness on 3D skeleton data estimated from large-scale RGB videos. Our codes are open at \href{https://github.com/Kali-Hac/SM-SGE}{https://github.com/Kali-Hac/SM-SGE.}
\end{abstract}





\maketitle

\section{Introduction}
Person re-identification (Re-ID) aims to retrieve the same individual from a different view or scene, with great potential in authentication-related applications \cite{vezzani2013people}. Conventional studies \cite{wang2018learning,karianakis2018reinforced,haque2016recurrent} typically extract appearance-based features such as body texture and silhouettes from RGB or depth images to perform person Re-ID. Nevertheless, an important flaw of these methods is their vulnerability to illumination or appearance changes \cite{rao2021a2}. In contrast, skeleton-based models exploit 3D coordinates of key body joints to characterize human body and motion, which are usually robust to factors such as view and body shape changes \cite{han2017space}. Despite that skeleton data have been extensively studied in action and motion related tasks \cite{li2020dynamic}, it is still an open challenge to extract discriminative body and motion features with 3D skeletons for person Re-ID \cite{rao2020self}. In this sense, this work aims to
construct a systematic framework from three aspects to tackle the skeleton-based person Re-ID task.

\begin{figure}
    \centering
    \scalebox{0.36}{
    \includegraphics{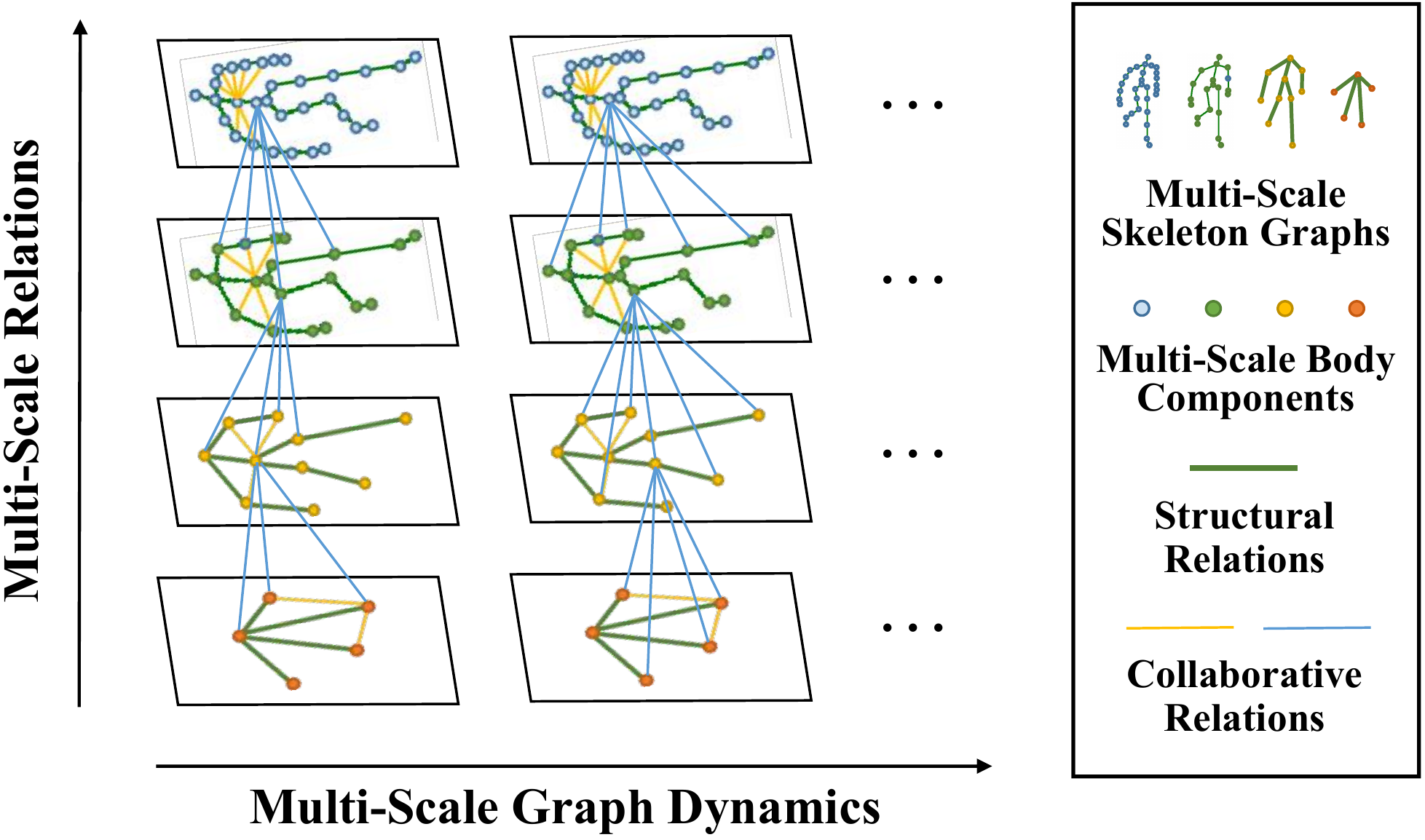} }
    \caption{SM-SGE exploits multi-scale skeleton graphs to model body structure and internal relations (structural and collaborative relations), and captures multi-scale graph dynamics to learn skeleton representations for person Re-ID.}
    \label{first_overview}
\end{figure}

(1) \textbf{Multi-scale skeleton graphs.} 
Most existing works \cite{barbosa2012re,munaro2014one,andersson2015person,pala2019enhanced}  construct skeleton descriptors to depict discriminative features of body structure and motion ($e.g.,$ anthropometric and gait attributes \cite{andersson2015person}) for person Re-ID. However, they typically extract these hand-craft features from skeletons with a single spatial scale and topology, which limits their ability to capture underlying structural information from different body partitions beyond body-joint level ($e.g.,$ limb-level components) \cite{li2020dynamic}. To fully mine latent structural features within body structure, it is beneficial to devise a systematic manner to represent skeletons at different levels. In this work, we model skeletons as \textit{multi-scale} graphs (see Fig. \ref{first_overview}) to learn coarse-to-fine grained body and motion features from 3D skeleton data.

(2) \textbf{Multi-scale relation learning.} 
In human walking, body components usually possess different internal relations, which could carry unique and recognizable patterns \cite{murray1964walking,winter2009biomechanics}. Recent works like \cite{rao2020self,liao2020model,rao2021a2} typically encode body-joint trajectory or pre-defined pose descriptors into a feature vector for skeleton representation learning, while they rarely explore the inherent relations between different body joints or components. For example, adjacent body joints ``knee'' and ``foot'' are strongly correlated in walking, while they enjoy different degrees of \textit{collaboration} with their corresponding limb-level component ``leg''. To capture such internal body relations in skeletal motion, it is highly desirable to design a framework to capture 
correlations between physically-connected body parts (referred as ``\textit{structural relations}'') and relations among all collaborative body components (referred as ``\textit{collaborative relations}'').

(3) \textbf{Multi-scale skeleton dynamics modeling.}  Existing 3D skeleton based Re-ID methods usually model skeleton dynamics from the trajectory of body joints \cite{rao2020self,rao2021a2} or the sequence of pre-defined joint features ($e.g.,$ pairwise joint distances and pose descriptors \cite{liao2020model}). 
Since these methods learn skeleton motion at a fixed scale of body joints, they lack the flexibility to capture motion patterns at various levels. For instance, they cannot explicitly model the movement or interaction of higher level limbs from joint trajectory, which might cause a loss of global motion features. Hence, it is important to devise a framework that can explicitly model skeleton dynamics at different scales to better capture body motion patterns.

To fulfill all above goals, this work \textit{for the first time} proposes a \textbf{S}elf-supervised \textbf{M}ulti-scale \textbf{S}keleton \textbf{G}raph \textbf{E}ncoding (SM-SGE) framework that exploits coarse-to-fine skeleton graphs to model body-structure and motion features for person Re-ID. Specifically, we first construct \textit{multi-scale skeleton graphs} by spatially dividing each skeleton into body-component nodes of different granularities (shown in Fig. \ref{graphs}), which allows our framework to fully model body structure and capture skeleton features at various levels.
Second, motivated by the fact that human walking usually carries unique patterns \cite{murray1964walking}, which endow body components with different internal relations, we propose a \textbf{M}ulti-scale \textbf{G}raph \textbf{R}elation \textbf{N}etwork (MGRN) to capture \textit{structural} and \textit{collaborative} relations among body components in multi-scale skeleton graphs. MGRN exploits structural relations between adjacent body-component nodes to aggregate key correlative features for better node representations, and meanwhile incorporates collaborative relations among nodes of different scales into graph encoding process to enhance global pattern learning. Finally, we propose a novel \textbf{M}ulti-scale \textbf{S}keleton \textbf{R}econstruction (MSR) mechanism with two concurrent pretext tasks, namely \textit{skeleton subsequence reconstruction} task and \textit{cross-scale skeleton inference} task, to enable our framework to capture skeleton dynamics and latent high-level semantics ($e.g.,$ body part correspondence, sequence order) from \textit{unlabeled} skeleton graph representations. The graph features of all scales learned from the proposed framework are then combined as the final skeleton representation to perform the downstream task of person Re-ID.


The proposed SM-SGE framework enjoys three main advantages: First, it seamlessly unifies the learning of multi-scale skeleton graphs into a systematic framework, which enables us to model body structure, component relations, and motion patterns of skeletons at different levels. Second, unlike most existing skeleton-based methods that require manual annotation ($e.g.,$ ID labels) for representation learning, our framework is able to learn an effective representation for unlabeled skeletons, which can be directly applied to skeleton-based tasks such as person Re-ID.
Last, our framework is also effective with the 3D skeleton data estimated from RGB videos \cite{yu2006framework}, thus it can be potentially applied to RGB-based datasets under general settings. 
In summary, our main contributions include:
\begin{itemize}
    \item We devise multi-scale graphs to fully model 3D skeletons, and propose a novel self-supervised multi-scale skeleton graph encoding (SM-SGE) framework to learn an effective representation from unlabeled skeletons for person Re-ID.
    
    \item We propose the multi-scale graph relation network (MGRN) to learn both structural and collaborative relations of body-component nodes, so as to aggregate crucial correlative features of nodes and capture richer pattern information.
    
    \item We propose the multi-scale skeleton reconstruction (MSR) mechanism to enable the framework to encode graph dynamics and high-level semantics from unlabeled skeletons.
    
    \item Extensive experiments show that SM-SGE outperforms most state-of-the-art skeleton-based methods on three person Re-ID benchmarks, and it can achieve highly competitive performance on skeletons estimated from large-scale RGB videos.
\end{itemize}

\section{Related Works}
This section briefly reviews existing skeleton-based person Re-ID methods using hand-crafted features, supervised learning or self-supervised learning. We also introduce depth-based and multi-modal Re-ID methods that are related to skeleton-based models.

\textbf{Hand-crafted and Supervised Re-ID Methods with Skeleton Data.} Most existing works extract hand-crafted skeleton descriptors to depict certain geometric, anthropometric, and gait attributes of human body. \cite{barbosa2012re} computes 7 Euclidean distances between the floor plane and joint or joint pairs to construct a distance matrix, which is learned to match gallery individuals with a quasi-exhaustive strategy.
 \cite{munaro2014one} and \cite{pala2019enhanced} further extend them to 13 ($D^{13}$) and 16 skeleton descriptors ($D^{16}$) respectively, and leverage support vector machine (SVM), $k$-nearest neighbor (KNN) or Adaboost classifiers for Re-ID. As existing solutions that use skeleton features alone typically perform unsatisfactorily, other modalities such as 3D face descriptors \cite{pala2019enhanced} and 3D point clouds \cite{munaro20143d} are often used to boost the performance. A few recent studies resort to supervised deep learning models to learn discriminative skeleton representation: \cite{haque2016recurrent} utilizes long short-term memory (LSTM) \cite{hochreiter1997long} to encode temporal dynamics of pairwise joint distance to perform person Re-ID; Liao \textit{et al.} \cite{liao2020model}
 propose PoseGait model, which learns 81 hand-crafted pose features of 3D skeleton data with deep convolutional neural networks (CNN) for gait-based human recognition. 
 
 \textbf{Self-supervised Skeleton-based Re-ID Methods.} Recently, Rao \textit{et al.} \cite{rao2020self} devise a self-supervised attention-based gait encoding model with multi-layer LSTM to encode gait features from unlabeled skeleton sequences for person Re-ID. The latest self-supervised study \cite{rao2021a2} further proposes a locality-awareness approach that combines various pretext tasks ($e.g.,$ reverse sequential reconstruction) and contrastive learning scheme to enhance self-supervised gait representation learning for the person Re-ID task.

\textbf{Depth-based and Multi-modal Re-ID Methods.} 
Depth-based methods exploit depth-image sequences to extract human shapes, silhouettes or gait features for person Re-ID. \cite{sivapalan2011gait} proposes Gait Energy Volume (GEV) algorithm, which extends Gait Energy Image (GEI) \cite{chunli2010behavior} to 3D domain, to learn depth features for human recognition. \cite{munaro2014one} devises a depth-based point cloud matching (PCM) method to match multi-view 3D point cloud sets to discriminate different individuals. In  \cite{haque2016recurrent},
Haque \textit{et al.} leverage 3D LSTM and 3D CNN \cite{boureau2010theoretical} to learn motion dynamics from 3D point clouds for person Re-ID.
As to multi-modal methods, skeleton information and RGB or depth features ($e.g.,$ depth shape features \cite{munaro20143d,wu2017robust,hasan2016long}) are usually combined to enhance Re-ID performance. In \cite{karianakis2018reinforced}, Karianakis \textit{et al.} propose an RGB-to-depth transferred CNN-LSTM model with reinforced temporal attention (RTA) for the person Re-ID task.

\section{The proposed Framework}
\begin{figure*}[t]
    \centering
    \scalebox{0.483}{
    \includegraphics{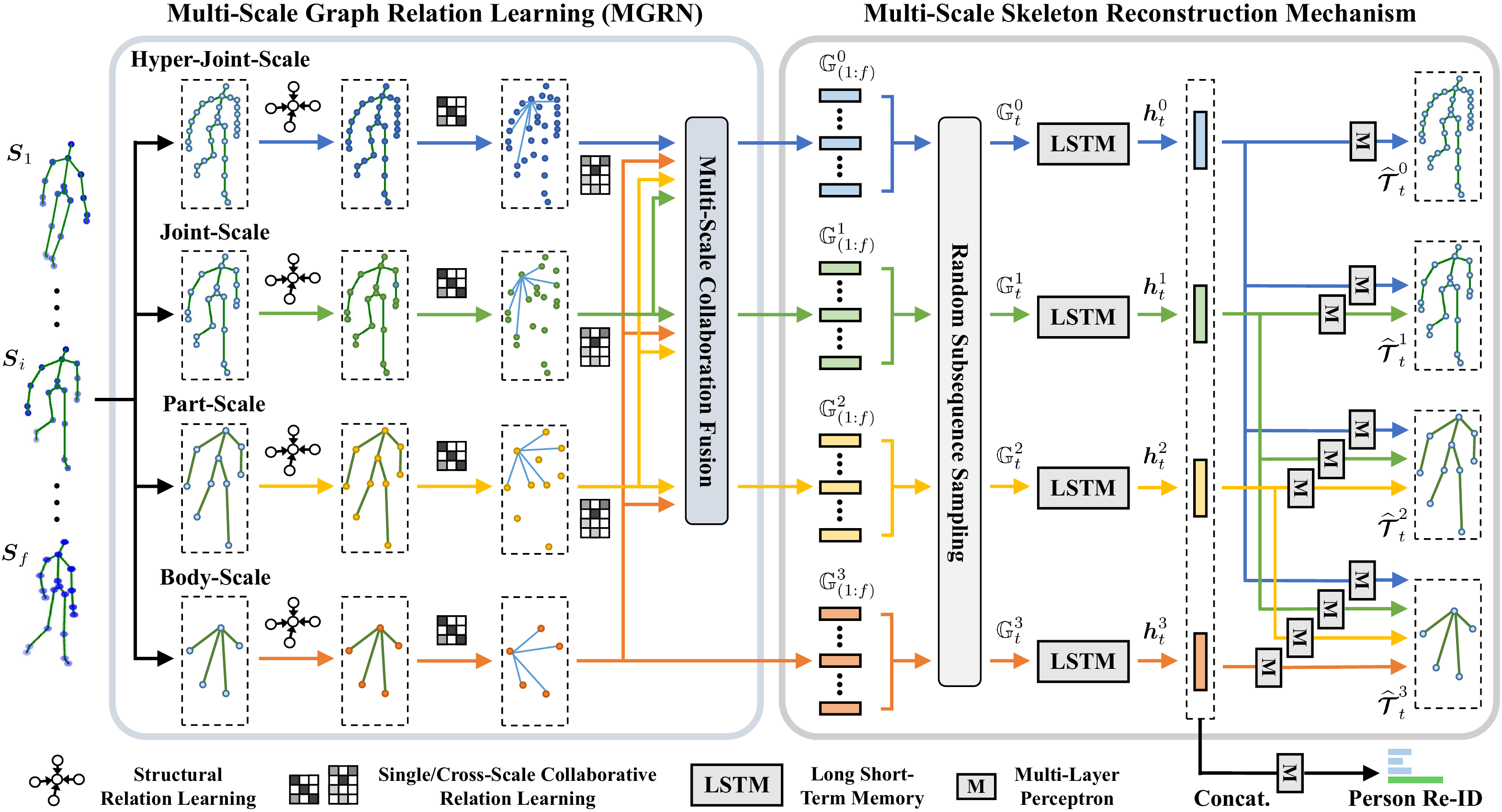}
    }
    \caption{Schematic diagram of SM-SGE: First, we construct graphs of four scales for each skeleton in a sequence $\boldsymbol{S}_{(1:f)}$. Second, we exploit multi-scale graph relation network (MGRN) to capture structural relations of neighbor nodes to aggregate crucial structural features for node representations, and compute both single-scale and cross-scale collaborative relations among body-component nodes, which are exploited to fuse collaborative node features across scales.
    Then, we utilize LSTM to encode the fused $m^{th}$ scale graph representation ($\mathbb{G}^{m}_{t}$) of the $t^{th}$ skeleton in the subsequence, which is randomly sampled from $\boldsymbol{S}_{(1:f)}$ (corresponding to graph features $\mathbb{G}^{m}_{(1:f)}$), into encoded graph state ($\boldsymbol{h}^{m}_{t}$) to capture graph dynamics and reconstruct skeletons ($\widehat{\boldsymbol{\mathcal{T}}}^{m}_{t}$) across scales. Finally, the learned encoded graph states of all scales are concatenated and fed into MLP for person Re-ID.}
    \label{model}
\end{figure*}

Suppose that a 3D skeleton sequence with $f$ consecutive skeletons is $\boldsymbol{S}_{(1:f)}\!=\!(\boldsymbol{S}_1,\cdots,\boldsymbol{S}_{f})\in \mathbb{R}^{f \times J \times D}$, where $\boldsymbol{S}_{t}\in \mathbb{R}^{J \times D}$ denotes the $t^{th}$ skeleton with three-dimensional coordinates  ($D\!=\!3$) of $J$ body joints. 
 $\Phi=\left\{\boldsymbol{S}^{(i)}_{(1:f)}\right\}_{i=1}^{N}$ represents the training set that contains $N$ skeleton sequences collected from different persons and views. 
Each skeleton sequence $\boldsymbol{S}^{(i)}_{(1:f)}$ corresponds to an ID label $y_{i}$, where $y_{i}\in \{1, \cdots, C\}$ and $C$ is the number of different persons. The goal of SM-SGE framework is to learn a latent discriminative representation $\textbf{H}$ from skeleton sequences $\boldsymbol{S}$ without using any label. Then, we evaluate the effectiveness of learned skeleton representation ($\textbf{H}$) on the downstream task of person Re-ID: \textit{Frozen} $\textbf{H}$ and corresponding ID labels are used to train a multi-layer perceptron (MLP) for person Re-ID (note that the learned features $\textbf{H}$ are NOT tuned at this training stage). The overview of proposed SM-SGE framework is shown in Fig. \ref{model}, and we present each technical component as below.

\subsection{Multi-Scale Skeleton Graph Construction}
\label{graph_construct}
\begin{figure}[t]
    \centering
    \scalebox{0.5}{
    \includegraphics{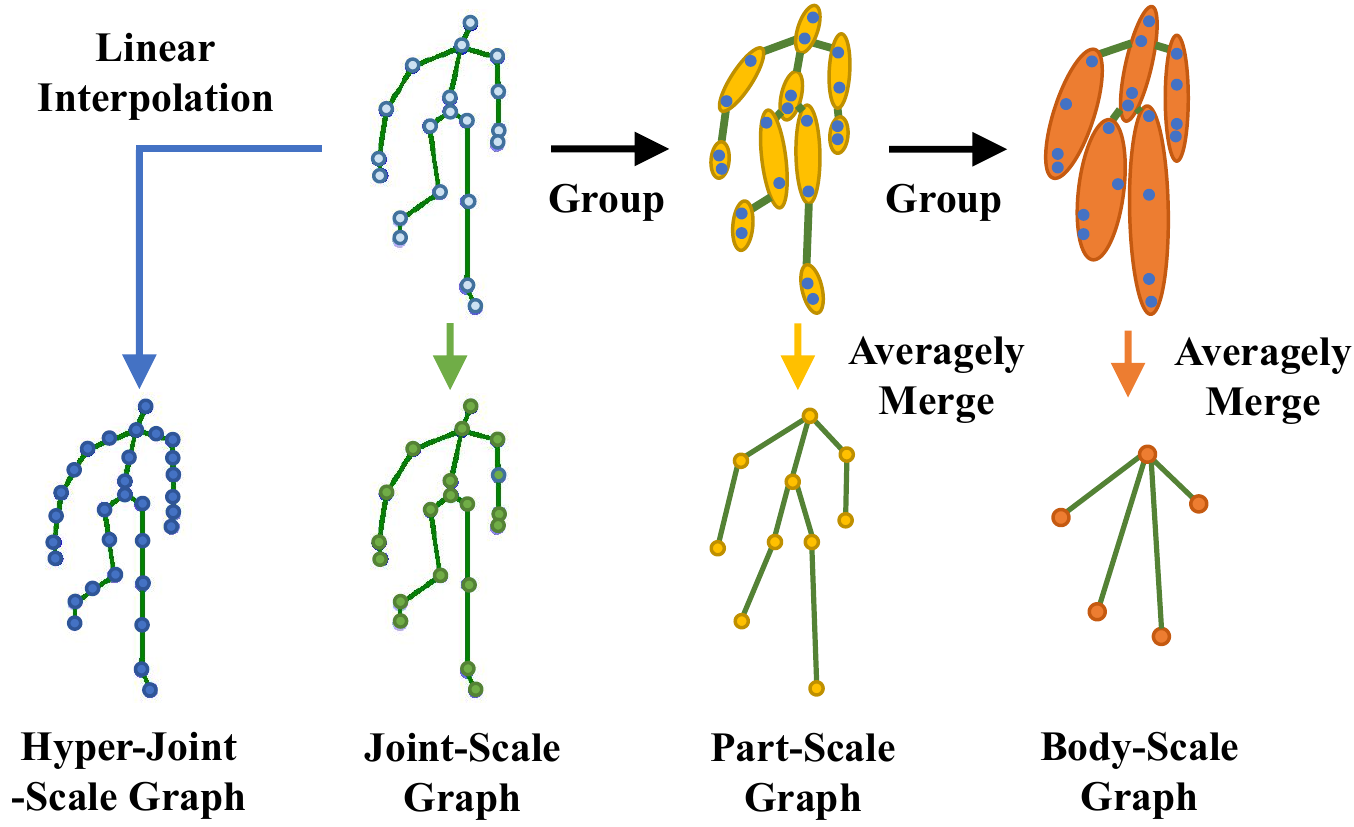}
    }
    \caption{Four graph scales for a skeleton with 20 body joints. We divide body into 10 and 5 parts to build part-scale and body-scale graphs, and merge internal joints into nodes.}
    \label{graphs}
\end{figure}

Human body can be segmented into functional components with diverse granularities ($e.g.,$ knee joint, thigh part, leg limb), each of which typically carries different geometric or anthropometric attributes of body \cite{winter2009biomechanics}. Inspired by this fact, we regard body joints as the basic components, and merge spatially nearby groups of joints to be a higher level body-component node at the center of their positions. 
As shown in Fig. \ref{graphs},
we first construct skeleton graphs at three scales, namely \textit{joint-scale}, \textit{part-scale}, and \textit{body-scale} graphs (denoted as $\mathcal{G}^1, \mathcal{G}^2, \mathcal{G}^3$) for each skeleton $\boldsymbol{S}$. Besides, to encourage our model to capture coarse-to-fine skeleton features more systematically, we also build a \textit{hyper-joint-scale} graph (denoted as $\mathcal{G}^0$) based on a denser body-limb representation \cite{liu2018recognizing}, which is constructed by linearly interpolating nodes between adjacent nodes in the joint-scale graph.
Each graph $\mathcal{G}^{m}(\mathcal{V}^{m}, \mathcal{E}^{m})$  ($m\in\{0,1,2,3\}$) consists of nodes $\mathcal{V}^{m}=\{\boldsymbol{v}^{m}_{1}, \boldsymbol{v}^{m}_{2}, \cdots,\boldsymbol{v}^{m}_{n_m}\}$ ($\boldsymbol{v}^{m}_{i}\in\mathbb{R}^{D}$, $i\in\{1,\cdots,n_m\}$) and edges $\mathcal{E}^{m}=\{e^{m}_{i,j}\ | \boldsymbol{v}^{m}_{i}, \boldsymbol{v}^{m}_{j}\!\in\!\mathcal{V}^{m}\} $ ($e^{m}_{i,j}\in\mathbb{R}$). Here $\mathcal{V}^{m}$, $\mathcal{E}^{m}$ denote the set of nodes corresponding to different body components and the set of their structural relations respectively, and $n_m$ is the number of nodes in the $m^{th}$ scale graph $\mathcal{G}^{m}$. 
We use $\mathbf{A}^{m} \in \mathbb{R}^{n_m \times n_m}$ to represent a graph's adjacency matrix, where each element $\mathbf{A}^{m}_{i,j}$ is defined as the \textit{normalized} structural relation between adjacent nodes $i$ and $j$, and satisfy: $\sum_{j \in \mathcal{N}_{i}}\mathbf{A}^{m}_{i,j}=1$, where $\mathcal{N}_{i}$ denotes indices for neighbor nodes of node $i$ in $\mathcal{G}^{m}$. During training of SM-SGE, $\mathbf{A}^{m}$ is adaptively learned to capture flexible structural relations.

\subsection{Multi-Scale Graph Relation Network}
\label{MGRN}
Different body parts typically possess internal relations at the physical or kinematic level, which could be exploited to mine rich body-structure features and patterns of motion \cite{aggarwal1998nonrigid}. Motivated by this fact, we propose to learn relations of body components from two aspects:
\textbf{(1)} \textit{Structural relations}: Structurally-connected body components usually enjoy a higher motion correlation than distant pairs. Thus, to better represent each body-component node, when encoding skeleton graph features, it is crucial to capture structural relations of neighbor nodes to aggregate the most correlative spatial features. \textbf{(2)} \textit{Collaborative relations}: Human motion like walking is often performed with several action-related body components, which collaborate together in a relatively stable pattern ($i.e.,$ gait patterns) \cite{murray1964walking,rao2020self}. It is therefore beneficial to learn the inherent collaborative relations among different body components to mine more global pattern information from skeletons. To achieve above goals, we propose the \textit{Multi-scale Graph Relation Network (MGRN)} with structural and collaborative relation learning as below.


\textbf{Structural Relation Learning.} Given the $m^{th}$ scale graph $\mathcal{G}^{m}$  of a skeleton, 
MGRN first computes the structural relation $e^{m}_{i, j}$ between \textit{adjacent} nodes ${\boldsymbol{v}}^{m}_{i}$ and ${\boldsymbol{v}}^{m}_{j}$ in $\mathcal{G}^{m}$ as follows:
\begin{equation}
\label{eq_1}
e^{m}_{i,j}=\sigma\!\left({\left(\mathbf{W}^{m}_{r}\right)}^{\mathsf{T}}\left[\mathbf{W}^{m}_{v} {\boldsymbol{v}}^{m}_{i} \| \mathbf{W}^{m}_{v} {\boldsymbol{v}}^{m}_{j}\right]\right)
\end{equation}
where $\mathbf{W}^{m}_{v}\in \mathbb{R}^{D_{t}\times D}$ is the weight matrix that maps the $m^{th}$ scale node $\boldsymbol{v}^{m}_{i}\in \mathbb{R}^{D}$ into a higher level feature space $ \mathbb{R}^{D_t}$, $\mathbf{W}^{m}_{r}\in \mathbb{R}^{2D_{t}}$ denotes a learnable weight matrix for relation learning at $m^{th}$ scale,  $\|$ indicates the feature concatenation of two nodes, and $\sigma(\cdot)$ is a non-linear activation function. Then, to learn flexible structural relations to focus on more correlative nodes, we normalize relations with a temperature-based $\operatorname{softmax}$ ($\operatorname{T-softmax}$) function \cite{hinton2015distilling} as follows:
\begin{equation}
\label{eq_2}
\mathbf{A}^{m}_{i,j}= \operatorname{T-softmax}_{j}\left(e^{m}_{i,j} \right) = \frac{\exp \left(e^{m}_{i,j} / T_1 \right)}{\sum_{k \in \mathcal{N}_{i}} \exp \left(e^{m}_{i,k} / T_1 \right)}
\end{equation}
where $T_1$ denotes the temperature that is normally set to $1$ in the $\operatorname{softmax}$ function, while higher value of $T_1$ produces a softer relation distributed over nodes and retains more similar relation information. Here $\mathcal{N}_{i}$ denotes neighbor nodes (including $i$) of node $i$ in graph. 

\begin{figure}[t]
    \centering
    \scalebox{0.55}{
    \includegraphics{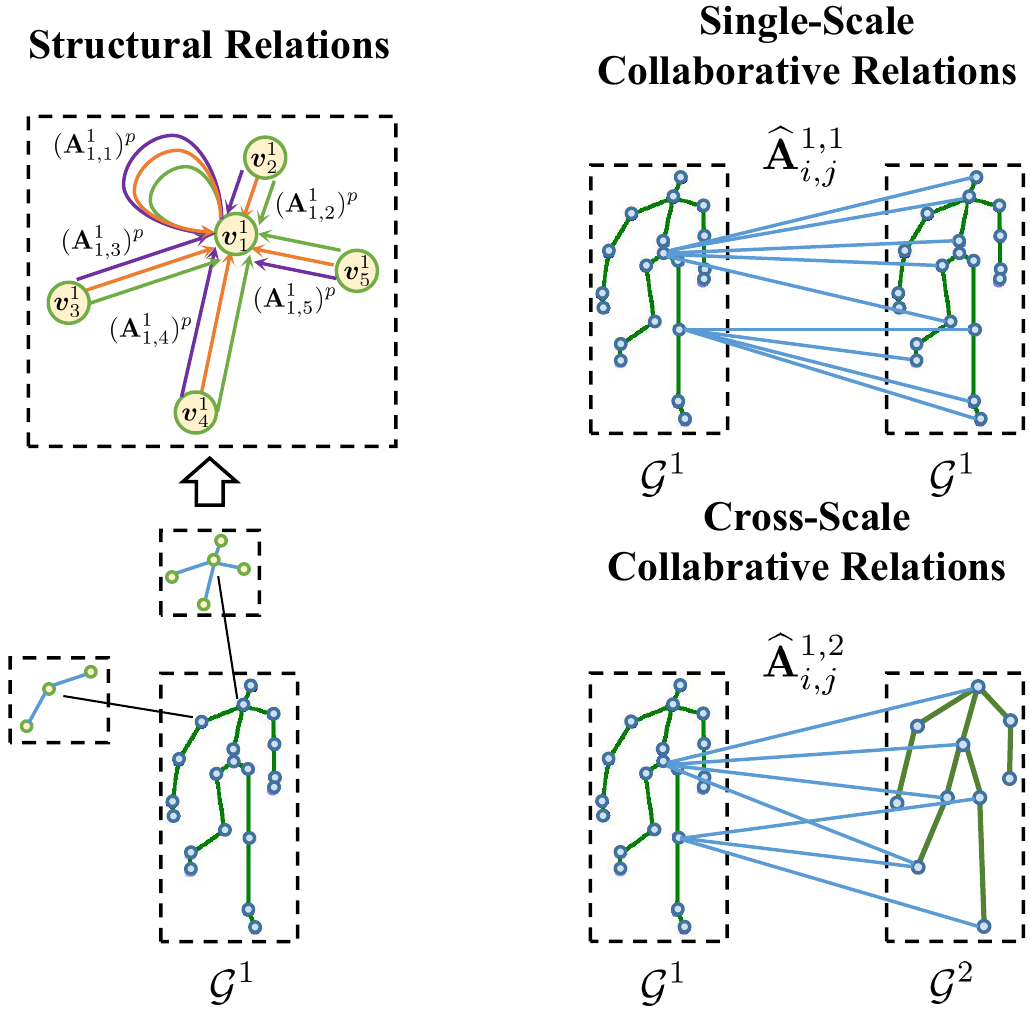}
    }
    \caption{Examples for three types of relations: (1) Structural relations in $\mathcal{G}^{1}$. (2) Single-scale collaborative relations in $\mathcal{G}^{1}$. (3) Cross-scale collaborative relations between $\mathcal{G}^{1}$ and $\mathcal{G}^{2}$.}
    \label{relation_learn}
\end{figure}

To aggregate features of most relevant nodes to represent the node $i$, we exploit normalized structural relations $\mathbf{A}^{m}_{i,j}$ to yield the representation $\boldsymbol{\overline{v}}^{m}_{i}\in \mathbb{R}^{D_{t}}$ for node $i$ by: 
$
\boldsymbol{\overline{v}}^{m}_{i}=\sigma\left(\sum_{j \in \mathcal{N}_{i}} \mathbf{A}^{m}_{i,j} \mathbf{W}^{m}_{v} \boldsymbol{v}^{m}_{j}\right)
$. To sufficiently capture potential structural relations ($e.g.,$ motion correlation, position similarity), MGRN concurrently and independently learns $P$ different structural relation matrices $(\mathbf{A}^{m}_{i,j})^{p}$  ($p\in\{1,\cdots,P\}$) using the same computation (see Eq. \ref{eq_1}, Eq. \ref{eq_2}). In this way, MGRN can jointly capture structural relation information of nodes from different representation subspaces \cite{velickovic2018graph} based on $P$ learnable structural relation matrices (see Fig. \ref{relation_learn}).
We \textit{averagely aggregate} features learned by these matrices to represent each node:
\begin{equation}
\label{node_repre}
\boldsymbol{\widehat{v}}^{m}_{i}= \frac{1}{P} \sum^{P}_{p=1}\sigma\left( \sum_{j \in \mathcal{N}_{i}} (\mathbf{A}^{m}_{i,j})^{p}\ (\mathbf{W}^{m}_{v})^{p}\  \boldsymbol{v}^{m}_{j}\right)
\end{equation}
where $\boldsymbol{\widehat{v}}^{m}_{i}\in\mathbb{R}^{D_{t}}$ denotes the $i^{th}$ node representation of $\mathcal{G}^{m}$ learned by $P$ structural relation matrices, $(\mathbf{A}^{m}_{i,j})^{p}\in\mathbb{R}$ represents the structural relation between node $i$ and $j$ computed by the $p^{th}$ structural relation matrix, and $(\mathbf{W}^{m}_{v})^p$ denotes the corresponding weight matrix to perform node feature mapping. 
Here we use \textit{average} rather than concatenation operation to reduce the feature dimension of nodes and allow for learning more structural relation matrices. 

\textbf{Collaborative Relation Learning.} Motivated by the fact that unique walking patterns could be represented by the dynamic cooperation among body joints or between different body components \cite{murray1964walking},  we expect our model to capture more discriminative patterns \textit{globally} by learning collaborative relations from two aspects: (1) \textit{single-scale} collaborative relations among nodes of the same scale, and (2) \textit{cross-scale} collaborative relations between a node and its spatially corresponding or motion-related \textit{higher} level body component. To this end, MGRN computes collaborative relation matrix
 $\mathbf{\widehat{A}}^{a, b} \in \mathbb{R}^{n_a \times n_{b}}$ ($a, b\in \{0, 1, 2, 3\}, a\leq b$) between $a^{th}$ scale nodes $\mathcal{V}^{a}$ and $b^{th}$ scale nodes $\mathcal{V}^{b}$ as follows (shown in Fig. \ref{relation_learn} and Fig.  \ref{model}): 
\begin{equation}
\label{CR_matrices_compute}
\mathbf{\widehat{A}}^{a, b}_{i, j}\!=\!\operatorname{T-softmax}_{j}\left({\boldsymbol{\widehat{v}}^{a}_{i}}^{\top} \boldsymbol{\widehat{v}}^{b}_{j}\right)\!=\!\frac{\exp \left({\boldsymbol{\widehat{v}}^{a}_{i}}^{\top} \boldsymbol{\widehat{v}}^{b}_{j}/T_2\right)}{\sum^{n_{b}}_{k=1} \exp \left({\boldsymbol{\widehat{v}}^{a}_{i}}^{\top} \boldsymbol{\widehat{v}}^{b}_{k}/T_2\right)}
\end{equation}
where $\mathbf{\widehat{A}}^{a, b}_{i, j}$ represents the singe-scale (when $a\!=\!b$) and cross-scale (when $a\!\neq\!b$) collaborative relation between node $i$ in $\mathcal{G}^a$ and node $j$ in $\mathcal{G}^{b}$. 
Here MGRN computes the inner product of node feature representations, which aggregate key spatial information with structural relation learning (see Eq. \ref{node_repre}), to measure the degree of collaboration between two nodes. $T_2$ denotes the temperature to adjust the softness of relation learning (illustrated in Eq. \ref{eq_2}).

\textbf{Multi-scale Collaboration Fusion.}
To adaptively focus on key correlative features in body-component collaboration at different spatial levels to enhance global pattern learning, we propose the multi-scale collaboration fusion that exploits collaborative relations to fuse node features across scales. Each node representation ($\widehat{\boldsymbol{v}}_{i}^{a}$) in the $a^{th}$ scale graph is updated by the feature fusion of collaborative nodes ($\widehat{\boldsymbol{v}}_{j}^{b}$) learned from different graphs as below (see Fig. \ref{model}): 
\begin{equation}
\label{eq_fuse}
\widehat{\boldsymbol{v}}_{i}^{a} \ \leftarrow \ \widehat{\boldsymbol{v}}_{i}^{a}+ \lambda_{C} \sum^{3}_{b=a}\left(\sum^{n_{b}}_{j=1}\mathbf{\widehat{A}}^{a, b}_{i, j}\ \mathbf{W}^{a,b}_{C} \  \boldsymbol{\widehat{v}}^{b}_{j}\right) 
\end{equation}
where $\mathbf{W}^{a,b}_{C}\in\mathbb{R}^{D_{t}\times D_{t}}$ is a learnable weight matrix to integrate collaborative features of $b^{th}$ scale node ($\widehat{\boldsymbol{v}}_{j}^{b}$) into $a^{th}$ scale node representation ($\widehat{\boldsymbol{v}}_{i}^{a}$), $n_{b}$ represents the number of nodes in $b^{th}$ scale graph, and $\lambda_{C}$ is the fusion coefficient to fuse collaborative graph node features. We denote the  fused graph features of $m^{th}$ scale ($m\in\{0,1,2,3\}$) for a skeleton sequence $\boldsymbol{S}_{(1:f)}$ be $\mathbb{G}^{m}_{(1:f)}=(\mathbb{G}^{m}_{1},\cdots,\mathbb{G}^{m}_{f})\in\mathbb{R}^{f\times n_m \times D_{t}}$.
Note that the multi-scale collaboration fusion does NOT directly fuse graph features of all scales into a representation. Instead, graph representations of each individual scale is retained (shown in Fig. \ref{model}) to encourage our model to capture skeleton dynamics and pattern information at different levels.


\subsection{Multi-Scale Skeleton Reconstruction Mechanism}
\label{MSR}
  To enable SM-SGE to encode multi-scale graph dynamics of \textit{unlabeled} skeletons, we propose a \textit{self-supervised} Multi-scale Skeleton Reconstruction (MSR) mechanism to simultaneously capture skeleton graph dynamics and high-level semantics ($e.g.,$ skeleton order in the subsequence, cross-scale component correspondence) from different scales of graphs. 
Unlike the \textit{plain} reconstruction that learns to reconstruct the whole sequence at a sole scale, the objective of MSR is combined with two concurrent pretext tasks as follows: 

\textbf{(1)} \textit{Skeleton subsequence reconstruction} task, which reconstructs multiple skeleton subsequences based on their graph representations. In particular, MSR aims to reconstruct target multi-scale skeletons ($\boldsymbol{\mathcal{T}}^m$) corresponding to multi-scale graphs ($\mathcal{G}^m$) in subsequences, instead of reconstructing the original subsequences (for clarity, we use the vector $\boldsymbol{\mathcal{T}}^m_{t}\in\mathbb{R}^{n_m\times D}$ to represent all node positions in the $m^{th}$ scale graph of the $t^{th}$ skeleton in the subsequence).

\textbf{(2)} \textit{Cross-scale skeleton inference} task that exploits \textit{fine} skeleton graph representations to infer 3D positions of coarser body components. For instance, we propose to use joint-scale graph representations ($\mathbb{G}^{1}_{1},\cdots,\mathbb{G}^{1}_{f}$), which may contain richer spatial information with denser nodes, to infer nodes of body-scale skeletons ($\boldsymbol{\mathcal{T}}^3_{1},\cdots \boldsymbol{\mathcal{T}}^3_{f}$). It can also be viewed as a \textit{cross-scale reconstruction} task to reconstruct different scale nodes with the same skeleton graph. 

To \textit{simultaneously} achieve above two pretext tasks, we first sample $k$-length subsequences $\boldsymbol{\mathcal{T}}_{(1:k)}\!=\!(\boldsymbol{\mathcal{T}}_{1}, \cdots, \boldsymbol{\mathcal{T}}_{k}) \in  \mathbb{R}^{k\times J\times D}$ by randomly discarding $(f-k)$ skeletons from the input sequence $\boldsymbol{S}_{(1:f)}$. To exploit more potential samples for training, the random sampling process is repeated for $r$ rounds and each round covers all possible lengths from $1$ to $f-1$. Second, given an sampled skeleton subsequence $\boldsymbol{\mathcal{T}}_{(1:k)}$, the MGRN encodes its corresponding skeleton graphs of each scale into fused graph features $\mathbb{G}^{m}_{(1:k)}=(\mathbb{G}^{m}_{1},\cdots,\mathbb{G}^{m}_{k})$ (see Eq. \ref{eq_1}-\ref{eq_fuse}). Then, we leverage an LSTM to integrate the temporal dynamics of graphs \textit{at each scale} into effective representations: LSTM
encodes each skeleton graph representation $\mathbb{G}^{\text{m}}_{t}$ and the previous step's latent state $\boldsymbol{h}^{m}_{t-1}$ (if existed), which provides the temporal context information of $m^{th}$ scale graph representations, into the current latent state $\boldsymbol{h}^{m}_{t}$ ($t\in\{1, \cdots,k\}$):
\begin{equation}
\label{LSTM}
\boldsymbol{h}^{m}_{t}=\left\{\begin{array}{ll}
\phi_{m}\left(\mathbb{G}^{\text{m}}_{1}\right) & \text { if } \quad t=1 \\
\phi_{m}\left(\boldsymbol{h}^{m}_{t-1}, \mathbb{G}^{\text{m}}_{t}\right) & \text { if } \quad 1<t\leq k
\end{array}\right.
\end{equation}
where $\boldsymbol{h}^{m}_{t}\in \mathbb{R}^{D_h}$, $\phi_{m}(\cdot)$ denotes the LSTM encoder, which aims to capture long-term dynamics of graph representations at the $m^{th}$ scale.
$\boldsymbol{h}^{m}_{1}, \cdots, \boldsymbol{h}^{m}_{k}$ are \textit{encoded graph states} that contain crucial temporal encoding information of $m^{th}$ scale graph representations from time $1$ to $k$. Last, we exploits encoded graph states at the $a^{th}$ scale to reconstruct the target skeleton at the $b^{th}$ scale as follows:
\begin{equation}
\label{f_recon}
f(\boldsymbol{h}^{a}_{i})=\widehat{\boldsymbol{\mathcal{T}}}^{b}_{i}
\end{equation}
where $\widehat{\boldsymbol{\mathcal{T}}}^{b}_{i}\in \mathbb{R}^{n_b \times D}$ is the reconstructed $i^{th}$ skeleton: When $a=b$, Eq. \ref{f_recon} is the plain skeleton reconstruction at the same scale, and $a < b$ indicates the cross-scale skeleton inference.
$f(\cdot)$ is a network function built by MLP, where weights are NOT shared between different scales, $i.e.,$ we train different individual MLPs for each skeleton reconstruction at the same or different scales (see Fig. \ref{model}). 

As we expect to capture graph dynamics and pattern features of skeletons at various scales, we employ the MSR mechanism with the above reconstruction objective on all scales of graphs. Formally, we define the objective function $\mathcal{L}^{a}_{S}$ for the self-supervision of MSR on the $a^{th}$ scale graphs, which minimizes $\ell_{1}$ loss between ground-truth skeletons of the $k$-length subsequence and reconstructed skeletons:
\begin{equation}
\label{recon_loss}
\mathcal{L}^{a}_{S} = \sum^{3}_{b=a}\sum^{k}_{i=1}\|\boldsymbol{\mathcal{T}}^{b}_{i}-\widehat{\boldsymbol{\mathcal{T}}}^{b}_{i}\|_{1} 
\end{equation}
where $\widehat{\boldsymbol{{\mathcal{T}}}}^{b}_{i}$ is the reconstructed $b^{th}$ scale skeleton based on $a^{th}$ scale encoded graph states (see Eq. \ref{f_recon}), and $\|\cdot\|_{1}$ denotes $\ell_{1}$ norm. The reason for using  $\ell_{1}$ loss is twofold: It gives sufficient gradients to positions with small losses to facilitate precise spatial reconstruction, and meanwhile can alleviate gradient explosion with stable gradients for large losses \cite{li2020dynamic}. 
It should be noted that our implementation actually optimizes Eq. \ref{recon_loss}
on \textit{each individual graph scale}, and the sum of reconstruction loss for all sampled subsequences is computed. By learning to reconstruct skeletons of the same scale and infer cross-scale body-component nodes dynamically ($i.e.,$ use varying subsequences), MSR encourages our framework to integrate crucial skeleton dynamics and high-level semantics into encoded graph states to achieve better person Re-ID performance (see Sec. \ref{discussion}). 

\subsection{The Entire Framework}
The computation flow of the entire framework during self-supervised learning can be summarized as: $\boldsymbol{S}\rightarrow$
$\mathcal{G}$ (Sec. \ref{graph_construct})
$\rightarrow\mathbb{G}$ (Sec. \ref{MGRN}) $\rightarrow\boldsymbol{h}$ (Eq. \ref{LSTM})
$\rightarrow 
\widehat{\boldsymbol{\mathcal{T}}}$ (Eq. \ref{f_recon}). The self-supervised loss $\mathcal{L}_{S}$ (see Eq. \ref{recon_loss}) is employed to train the SM-SGE framework to learn an effective skeleton representation from multi-scale skeleton graphs. For the downstream task of person Re-ID, we extract encoded graph states ($\boldsymbol{h}$) learned from the pre-trained framework, and exploit an MLP ($g(\cdot)$) to predict the sequence label. Specifically, for the $t^{th}$ skeleton in an input sequence, we concatenate its corresponding encoded graph states of four scales, namely $\textbf{H}_{t}=[\boldsymbol{h}^{0}_{t};\boldsymbol{h}^{1}_{t};\boldsymbol{h}^{2}_{t};\boldsymbol{h}^{3}_{t}] \in \mathbb{R}^{4 \times D_{h}}$ ($t\in\{1,\cdots, f\}$), as the $t^{th}$ skeleton-level representation of the sequence. Then, we train the MLP ($g(\cdot)$) with the \textit{frozen} $\textbf{H}_{t}$ and its label (note that $\textbf{H}_{t}$ is NOT tuned in this training stage). The ID prediction ($g\left(\textbf{H}_{t}\right)$) of each skeleton-level representation in a sequence is averaged to be the final sequence-level ID prediction $\hat{y}$. We employ the cross-entropy loss to train $g(\cdot)$ for person Re-ID.

\begin{table}[t]
\centering
\caption{Performance comparison with hand-crafted, supervised and self-supervised methods on IAS-A and IAS-B. $^{*}$ and $^{\dagger}$ denote depth-based and multi-modal methods respectively. Bold numbers refer to the best performer among skeleton-based methods. ``—'' indicates no published results.}
\label{IAS_results}
\scalebox{0.75}{
\setlength{\tabcolsep}{1.5mm}{
\begin{tabular}{lrrrr}
\specialrule{0.1em}{0.45pt}{0.45pt}   
                                          & \multicolumn{2}{c}{\textbf{IAS-A}}                                      & \multicolumn{2}{c}{\textbf{IAS-B}}                                      \\
                                          & \multicolumn{1}{c}{\textbf{Rank-1}} & \multicolumn{1}{c}{\textbf{nAUC}} & \multicolumn{1}{c}{\textbf{Rank-1}} & \multicolumn{1}{c}{\textbf{nAUC}} \\ \specialrule{0.1em}{0.45pt}{0.45pt}   
\textbf{Hand-Crafted and Supervised Methods}      & \multicolumn{1}{l}{}                & \multicolumn{1}{l}{}              & \multicolumn{1}{l}{}                & \multicolumn{1}{l}{}              \\ \specialrule{0.1em}{0.45pt}{0.45pt}   
$^{*}$Gait Energy Image \cite{chunli2010behavior}               & 25.6                                & 72.1                              & 15.9                                & 66.0                              \\
$^{*}$3D CNN + Average Pooling \cite{boureau2010theoretical}       & 33.4                                & 81.4                              & 39.1                                & 82.8                              \\
$^{*}$Gait Energy Volume \cite{sivapalan2011gait}             & 20.4                                & 66.2                              & 13.7                                & 64.8                              \\
$^{*}$3D LSTM \cite{haque2016recurrent}                        & 31.0                                & 77.6                              & 33.8                                & 78.0                              \\
$^{\dagger}$PCM + Skeleton \cite{munaro20143d}               & 27.3                                & —                                 & 81.8                                & —                                 \\
$^{\dagger}$DVCov + SKL \cite{wu2017robust}                    & 46.6                                & —                                 & 45.9                                & —                                 \\
$^{\dagger}$ED + SKL \cite{wu2017robust}                      & 52.3                                & —                                 & 63.3                                & —                                 \\
$D^{13}$ descriptors + KNN \cite{munaro2014one}          & 33.8                                & 63.6                              & 40.5                                & 71.1                              \\
Single-layer LSTM \cite{haque2016recurrent}              & 20.0                                & 65.9                              & 19.1                                & 68.4                              \\
Multi-layer LSTM \cite{zheng2019relational}               & 34.4                                & 72.1                              & 30.9                                & 71.9                              \\
$D^{16}$ descriptors + Adaboost \cite{pala2019enhanced}     & 27.4                                & 65.5                              & 39.2                                & 78.2                              \\
PostGait \cite{liao2020model}                      & 41.4                                & 79.9                              & 37.1                                & 74.8                              \\ \specialrule{0.1em}{0.45pt}{0.45pt}   
\textbf{Self-Supervised Methods} & \multicolumn{1}{l}{}                & \multicolumn{1}{l}{}              & \multicolumn{1}{l}{}                & \multicolumn{1}{l}{}              \\ \specialrule{0.1em}{0.45pt}{0.45pt}   
Attention Gait Encodings \cite{rao2020self}       & 56.1                                & 81.7                              & 58.2                                & 85.3                              \\
SGELA \cite{rao2021a2}                         & \textbf{60.1}                       & 82.9                              & 62.5                                & 86.9                              \\
\textbf{SM-SGE (Ours)}                   & 59.4                                & \textbf{86.7}                     & \textbf{69.8}                       & \textbf{90.4}                     \\ \specialrule{0.1em}{0.2pt}{0.2pt}   
\end{tabular}
}
}
\end{table}

\section{Experiments}
\subsection{Experimental Settings}
\quad\textbf{Datasets:} We evaluate our framework on three public person Re-ID datasets that contain skeleton data (\textit{IAS-Lab} \cite{munaro2014feature}, \textit{KS20} \cite{nambiar2017context}, \textit{KGBD} \cite{andersson2015person}) and a large RGB video based multi-view dataset CASIA B \cite{yu2006framework}, which contain 11, 20, 164, and 124 different individuals respectively. We follow the frequently used evaluation setup in the literature \cite{rao2020self,haque2016recurrent}: For IAS-Lab, we use the full training set and two testing splits, IAS-A and IAS-B; For KGBD, since no training and testing splits are given, we randomly leave one skeleton video of each person for testing and use the remaining videos for training; For KS20, we randomly select one sequence from each viewpoint for testing and use the rest of skeleton sequences for training.

To evaluate the effectiveness of SM-SGE when 3D skeleton data are directly estimated from RGB videos rather than Kinect, we introduce a large-scale RGB video based dataset CASIA B \cite{yu2006framework}, 
and  exploit pre-trained pose estimation models \cite{chen20173d,cao2019openpose} to extract 3D skeletons from RGB videos (detailed in the Appendix). We evaluate our approach on each view ($0^{\circ}$, $18^{\circ}$, $36^{\circ}$, $54^{\circ}$, $72^{\circ}$, $90^{\circ}$, $108^{\circ}$, $126^{\circ}$, $144^{\circ}$, $162^{\circ}$, $180^{\circ}$) of CASIA B and use the adjacent views for training. 

\begin{table}[t]
\centering
\caption{Performance comparison on KGBD and KS20 datasets. Bold numbers refer to the best performer.}
\label{KS20_results}
\scalebox{0.75}{
\setlength{\tabcolsep}{1.5mm}{
\begin{tabular}{lrrrr}
\specialrule{0.1em}{0.45pt}{0.45pt}
\multicolumn{1}{c}{\textbf{}}             & \multicolumn{2}{c}{\textbf{KGBD}}                                       & \multicolumn{2}{c}{\textbf{KS20}}                                       \\
\textbf{}                                 & \multicolumn{1}{l}{\textbf{Rank-1}} & \multicolumn{1}{l}{\textbf{nAUC}} & \multicolumn{1}{l}{\textbf{Rank-1}} & \multicolumn{1}{l}{\textbf{nAUC}} \\ \specialrule{0.1em}{0.45pt}{0.45pt}
\textbf{Hand-Crafted and Supervised Methods}      & \multicolumn{1}{l}{}                & \multicolumn{1}{l}{}              & \multicolumn{1}{l}{}                & \multicolumn{1}{l}{}              \\ \specialrule{0.1em}{0.45pt}{0.45pt}
$D^{13}$ descriptors + KNN \cite{munaro2014one}          & 46.9                                & 90.0                              & 58.3                                & 78.0                              \\
Single-layer LSTM \cite{haque2016recurrent}              & 39.8                                & 87.2                              & 80.9                                & 92.3                              \\
Multi-layer LSTM \cite{zheng2019relational}              & 46.2                                & 89.8                              & 81.6                                & 94.2                              \\
$D^{16}$ descriptors + Adaboost \cite{pala2019enhanced}     & 69.9                                & 90.6                              & 59.8                                & 78.8                              \\
PostGait \cite{liao2020model}                       & 90.6                                & 97.8                              & 70.5                                & 94.0                              \\ \specialrule{0.1em}{0.45pt}{0.45pt}
\textbf{Self-Supervised Methods} & \multicolumn{1}{l}{}                & \multicolumn{1}{l}{}              & \multicolumn{1}{l}{}                & \multicolumn{1}{l}{}              \\ \specialrule{0.1em}{0.45pt}{0.45pt}
Attention Gait Encodings \cite{rao2020self}       & 87.7                                & 96.3                              & 86.5                                & 94.7                              \\
SGELA \cite{rao2021a2}                         & 86.9                                & 97.1                              & 86.9                                & 94.9                              \\
\textbf{SM-SGE (Ours)}                    & \textbf{99.5}                       & \textbf{99.6}                     & \textbf{87.5}                       & \textbf{95.8}                     \\ \specialrule{0.1em}{0.2pt}{0.2pt}
\end{tabular}
}
}
\end{table}

\textbf{Implementation Details:} The number of nodes in the hyper-joint-scale and joint-scale graph are $n_{0}\!=\!49$, $n_{1}\!=\!25$ in KS20, $n_{0}\!=\!27$, $n_{1}\!=\!14$ in CASIA B, and $n_{0}\!=\!39$, $n_{1}\!=\!20$ in IAS-Lab and KGBD datasets. For part-scale and body-scale graphs, the numbers of nodes are $n_2=10$ and $n_3=5$ for all datasets. On IAS-Lab, KS20 and KGBD datasets, the sequence length $f$ is empirically set to $6$, which achieves the best overall performance among different settings. For the largest dataset CASIA B with roughly estimated skeleton data from RGB frames, we set sequence length $f=20$ for training/testing.
The node feature dimension is $D_{t}=8$ and the number of structural relation matrices is $P=8$. The temperatures ($T_1,T_2$) for relation learning, the collaboration fusion coefficient ($\lambda_{C}$), and the number of sampling rounds ($r$) are empirically set to 1. MLP with one hidden layer is employed in SM-SGE. For MSR mechasnim, we use a 2-layer LSTM with $D_{h} = 256$ hidden units per layer. 
 We adopt Adam optimizer with learning rate $0.0025$ on IAS-Lab, KGBD and $0.0005$ on KS20, CASIA B to train the framework. 
 

\textbf{Evaluation Metrics:} 
Person Re-ID typically adopts a ``multi-shot'' manner that leverages predictions of multiple frames or a sequence-level representation to predict a sequence label. In this work, we compute both Rank-1 accuracy and nAUC (area under the cumulative matching curve (CMC) normalized by the number of ranks \cite{gray2008viewpoint}) to quantify multi-shot person Re-ID performance.

\subsection{Comparison with State-of-the-Art Methods}
In this section, we compare our approach with existing hand-crafted, supervised and self-supervised skeleton-based Re-ID methods on IAS-Lab (see Table \ref{IAS_results}), KS20 and KGBD (see Table \ref{KS20_results}). We also include classic depth-based methods and representative multi-modal methods as a reference. The comparison results are reported below:

\textbf{Comparison with Hand-crafted and Supervised Skeleton-based Methods:}
As shown in Table \ref{IAS_results} and Table \ref{KS20_results}, the proposed SM-SGE framework enjoys evident advantages over existing skeleton-based methods: First, our approach significantly outperforms two representative hand-crafted methods that extract anthropometric attributes of skeletons ($D^{13}$, $D^{16}$ descriptors \cite{munaro2014one,pala2019enhanced}) by $25.6\%$-$52.6\%$ Rank-1 accuracy and $9.0\%$-$23.1\%$ nAUC on different datasets. Second,
compared with state-of-the-art CNN-based (PoseGait \cite{liao2020model}) and LSTM-based models \cite{haque2016recurrent,zheng2019relational}, our self-supervised framework can achieve superior performance with a large margin (up to $59.7\%$ Rank-1 accuracy and $22.0\%$ nAUC) on all datasets. Besides, these supervised methods typically require massive labels and even extra hand-crafted features ($e.g.,$ PoseGait \cite{liao2020model} relies on 81 hand-crafted pose and motion features) for representation learning, while our framework is able to automatically model spatial and temporal features of \textit{unlabeled} skeleton graphs at various scales to learn a more effective skeleton representation for the person Re-ID task.

\textbf{Comparison with Self-supervised Skeleton-based Methods:} 
Our approach achieves a significant improvement ($0.6\%$-$12.6\%$ Rank-1 accuracy and $0.9\%$-$5.1\%$ nAUC) over existing state-of-the-art self-supervised methods on three out of four testing sets (IAS-B, KGBD, KS20). On IAS-A, despite both SGELA and our framework obtain a close Rank-1 accuracy, our approach gains a markedly higher nAUC ($3.8\%$) than SGELA, which suggests that our approach can achieve better overall Re-ID performance when retrieving persons from high to low ranking. Notably, the proposed SM-SGE outperforms existing self-supervised methods by more than $11.8\%$ Rank-1 accuracy on the largest skeleton-based dataset KGBD, which demonstrates the great potential of our approach on large-scale person Re-ID. 

\textbf{Comparison with Depth-based and Multi-modal methods:}
As reported in Table \ref{IAS_results}, our skeleton-based framework consistently performs better than classic depth-based methods (GEI \cite{chunli2010behavior}, GEV \cite{sivapalan2011gait}, 3D CNN \cite{boureau2010theoretical}, 3D LSTM \cite{haque2016recurrent}) by at least $26.0\%$ Rank-1 accuracy and $5.3\%$ nAUC on IAS-A and IAS-B. Compared with representative multi-modal methods, our approach is still the best performer in most cases. Interestingly, although the ``PCM + Skeleton'' method \cite{munaro20143d} that uses both skeletons and 3D point cloud matching attains the best Rank-1 accuracy on IAS-B, it is inferior to the SM-SGE by $32.1\%$ accuracy on IAS-A, which demonstrates that our approach is more effective under the setting with frequent shape and appearance changes (IAS-A). Considering that the proposed SM-SGE only requires 3D skeleton data as the input and can achieve more satisfactory performance on each dataset, it can be a promising solution to person Re-ID and other potential skeleton-related tasks.

\section{Further Analysis}
\label{discussion}
In this section, we first evaluate the performance of SM-SGE on skeleton data estimated from RGB videos in CASIA B. Then, we conduct ablation study to demonstrate the effectiveness of each component, and evaluate effects of different parameters on SM-SGE. Last, we visualize and analyze the learned collaborative relations.


\textbf{Evaluation with Model-estimated Skeletons.} We exploit pre-trained pose estimation models \cite{cao2019openpose,chen20173d} to extract 3D skeletons from RGB videos of CASIA B, and evaluate the performance of MS-SGE with the estimated skeleton data. 
We compare our framework with the state-of-the-art supervised method PoseGait \cite{liao2020model} under the same evaluation setup. As shown in Table \ref{CVE_comparison}, our approach significantly outperforms PoseGait by $1.1\%$-$40.4\%$ Rank-1 accuracy on all views of CASIA B.
Notably, SM-SGE obtains more stable performance than PoseGait on 7 consecutive views from $18^\circ$ to $126^\circ$, which shows the robustness of our framework to view-point variation. On the two most challenging views ($0^\circ$ and $180^\circ$), our approach can still perform better than PoseGait by more than $7.7\%$ Rank-1 accuracy.
These results demonstrate the effectiveness of our framework on skeleton data estimated from RGB videos, and also show the great potential of our approach to be applied to large RGB-based datasets under general settings ($e.g.,$ varying views).

\begin{table}[t]
\centering
\caption{Rank-1 accuracy on different views of CASIA B.
\label{CVE_comparison}}
\scalebox{0.75}{
\setlength{\tabcolsep}{1.05mm}{
\begin{tabular}{lccccccccccc}
\specialrule{0.1em}{0.45pt}{0.45pt} 
\multicolumn{1}{l}{\textbf{Methods}} & $\boldsymbol{0^{\circ}}$    & $\boldsymbol{18^{\circ}}$   & $\boldsymbol{36^{\circ}}$   & $\boldsymbol{54^{\circ}}$   & $\boldsymbol{72^{\circ}}$   & $\boldsymbol{90^{\circ}}$   & $\boldsymbol{108^{\circ}}$  & $\boldsymbol{126^{\circ}}$  & $\boldsymbol{144^{\circ}}$  & $\boldsymbol{162^{\circ}}$  & $\boldsymbol{180^{\circ}}$  \\ \specialrule{0.1em}{0.45pt}{0.45pt}
PoseGait [2020]                             & 10.7          & 37.4          & 52.5          & 28.3          & 24.3          & 18.9          & 23.5          & 17.2          & 23.6          & 18.8          & 4.3           \\
\textbf{SM-SGE (Ours)}                        & \textbf{18.4} & \textbf{50.8} & \textbf{53.6} & \textbf{40.9} & \textbf{51.2} & \textbf{59.3} & \textbf{52.3} & \textbf{53.9} & \textbf{30.2} & \textbf{28.8} & \textbf{13.6} \\ \specialrule{0.1em}{0.2pt}{0.2pt}
\end{tabular}
}
}
\end{table}

\begin{table}[t]
\centering
\caption{Rank-1 accuracy of SM-SGE with different components: Structural/collaborative relation (SR/CR) in MGRN, cross-scale skeleton inference (CSI) and skeleton subsequence reconstruction (SSR) in MSR. ``MG'' denotes exploiting multi-scale graphs rather than using joint-scale graphs. }
\label{ablation}
\scalebox{0.75}{
\setlength{\tabcolsep}{4.88mm}{
\begin{tabular}{ccccccc}
\specialrule{0.1em}{0.45pt}{0.45pt}
\multirow{2}{*}{\textbf{MG}} & \multicolumn{2}{c}{\textbf{MGRN}} & \multicolumn{2}{c}{\textbf{MSR}} & \multirow{2}{*}{\textbf{IAS-A}} & \multirow{2}{*}{\textbf{IAS-B}} \\
                             & \textbf{SR}     & \textbf{CR}     & \textbf{CSI}    & \textbf{SSR}   &                                 &                                 \\ \specialrule{0.1em}{0.45pt}{0.45pt}
                             &                 &                 &                 &                & 53.5                            & 61.8                            \\
                             & \checkmark               &                 &                 &                & 55.0                            & 65.8                            \\
 \checkmark                            &  \checkmark               &                 &                 &                & 56.1                            & 67.0                            \\
 \checkmark                            &  \checkmark               &                 & \checkmark               &                & 56.6                            & 67.8                            \\
 \checkmark                            &  \checkmark               &                 &  \checkmark               &  \checkmark              & 57.4                            & 68.9                            \\
 \checkmark                            &  \checkmark               &  \checkmark               &                 &                & 57.2                            & 67.7                            \\
 \checkmark                            &  \checkmark               &  \checkmark               &  \checkmark               &                & 57.9                            & 68.1                            \\
 \checkmark                            &  \checkmark               &  \checkmark               &  \checkmark               &  \checkmark              & \textbf{59.4}                   & \textbf{69.8}                   \\ \specialrule{0.1em}{0.2pt}{0.2pt}
\end{tabular}
}
}
\end{table}

\textbf{Ablation Study.} We evaluate the contribution of each component in our framework (here IAS-Lab is taken as an example) and report the results in Table \ref{ablation}. We use an LSTM with plain skeleton reconstruction as the baseline (see first row in Table \ref{ablation}). We can draw the following conclusions:
\textbf{(1)} Introducing multi-scale skeleton graphs (MG) consistently improves person Re-ID performance by at least $1.1\%$ Rank-1 accuracy, which justifies our claim that modeling skeletons as multi-scale graphs can facilitate learning richer body and motion features for person Re-ID.
\textbf{(2)} Exploiting structural relations (SR) between body components produces significant performance gain by $1.5\%$-$4.0\%$ Rank-1 accuracy compared with directly modeling body-joint trajectory (baseline), while combining collaborative relation (CR) learning further boosts the Re-ID performance by up to $2.0\%$ Rank-1 accuracy. These results demonstrate the effectiveness of multi-scale graph relation learning (MGRN) on capturing more discriminative body structural features and motion patterns for person Re-ID.
\textbf{(3)} The proposed MSR mechanism based on cross-scale skeleton inference and skeleton subsequence reconstruction pretext tasks evidently improves the model performance ($1.3\%$-$2.2\%$ Rank-1 accuracy) under different relation learning (SR or CR), which verifies our intuition that mining high-level semantics such as cross-scale body-component correspondence could encourage learning a more effective skeleton representation for the person Re-ID task. Other datasets also report similar results.

 \begin{figure}[t]
    \centering
    \scalebox{0.15}{\includegraphics[]{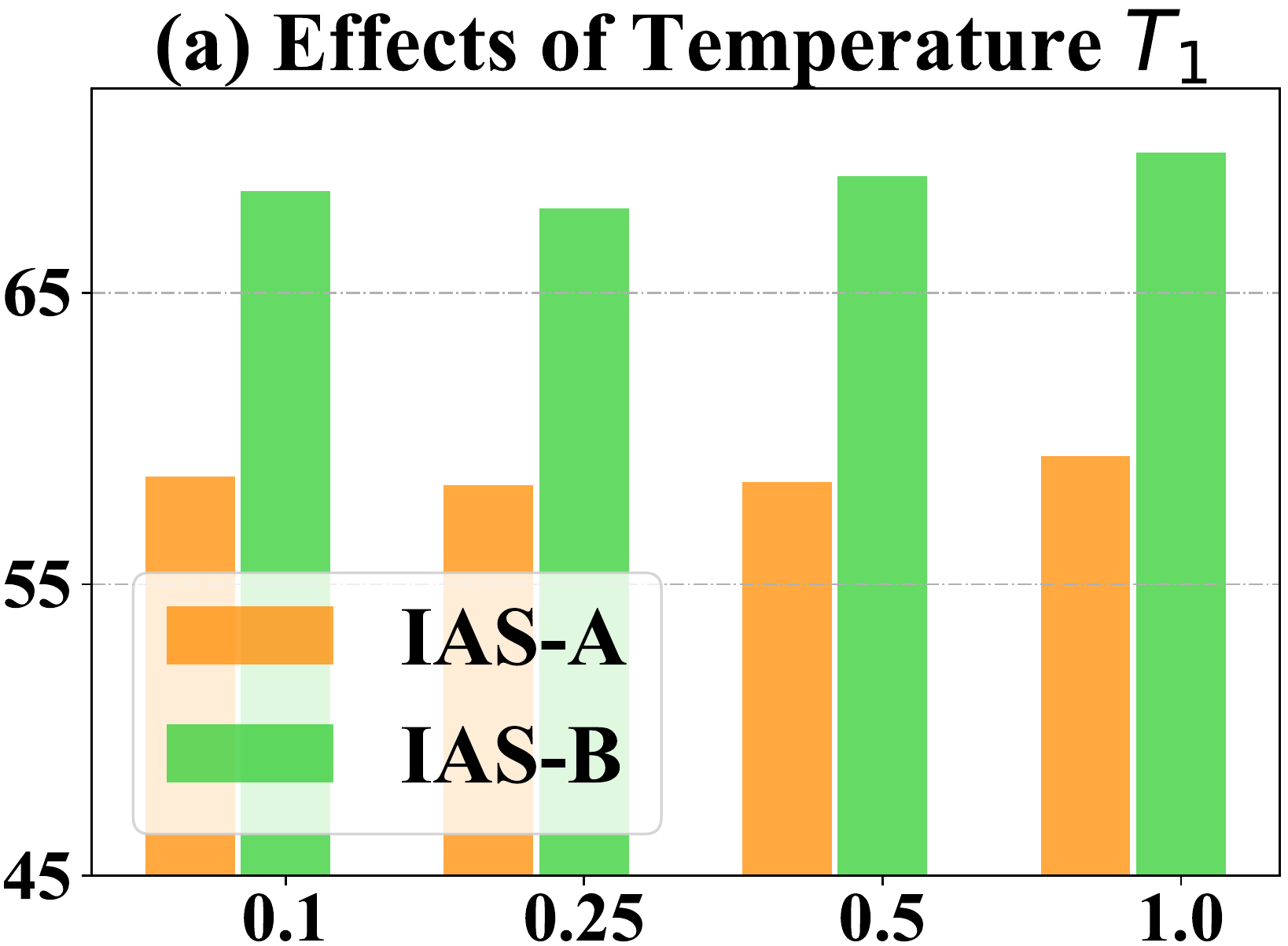}} \ \ \ 
  \scalebox{0.15}{\includegraphics[]{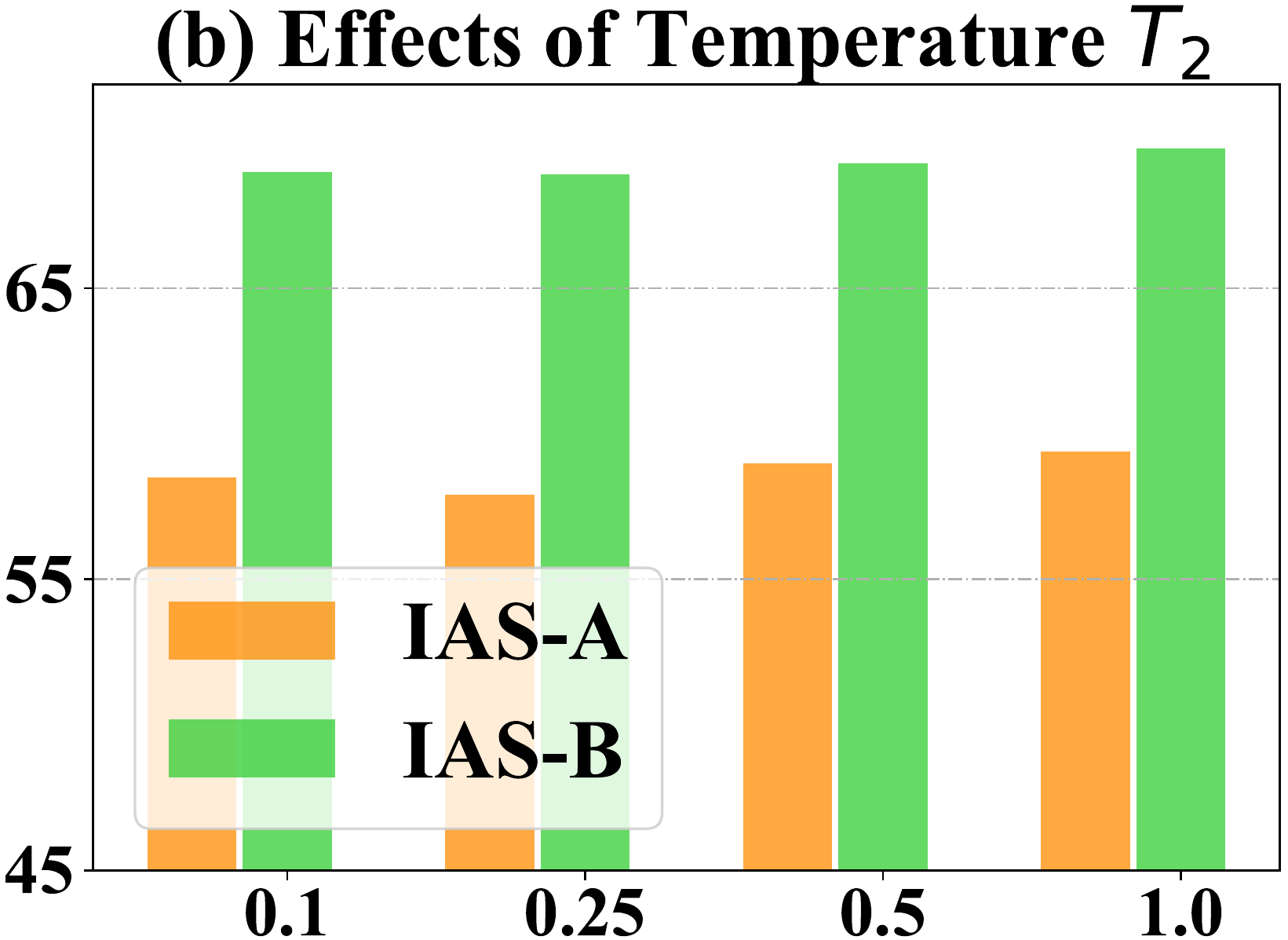}}\ \ \ 
  \scalebox{0.15}{\includegraphics[]{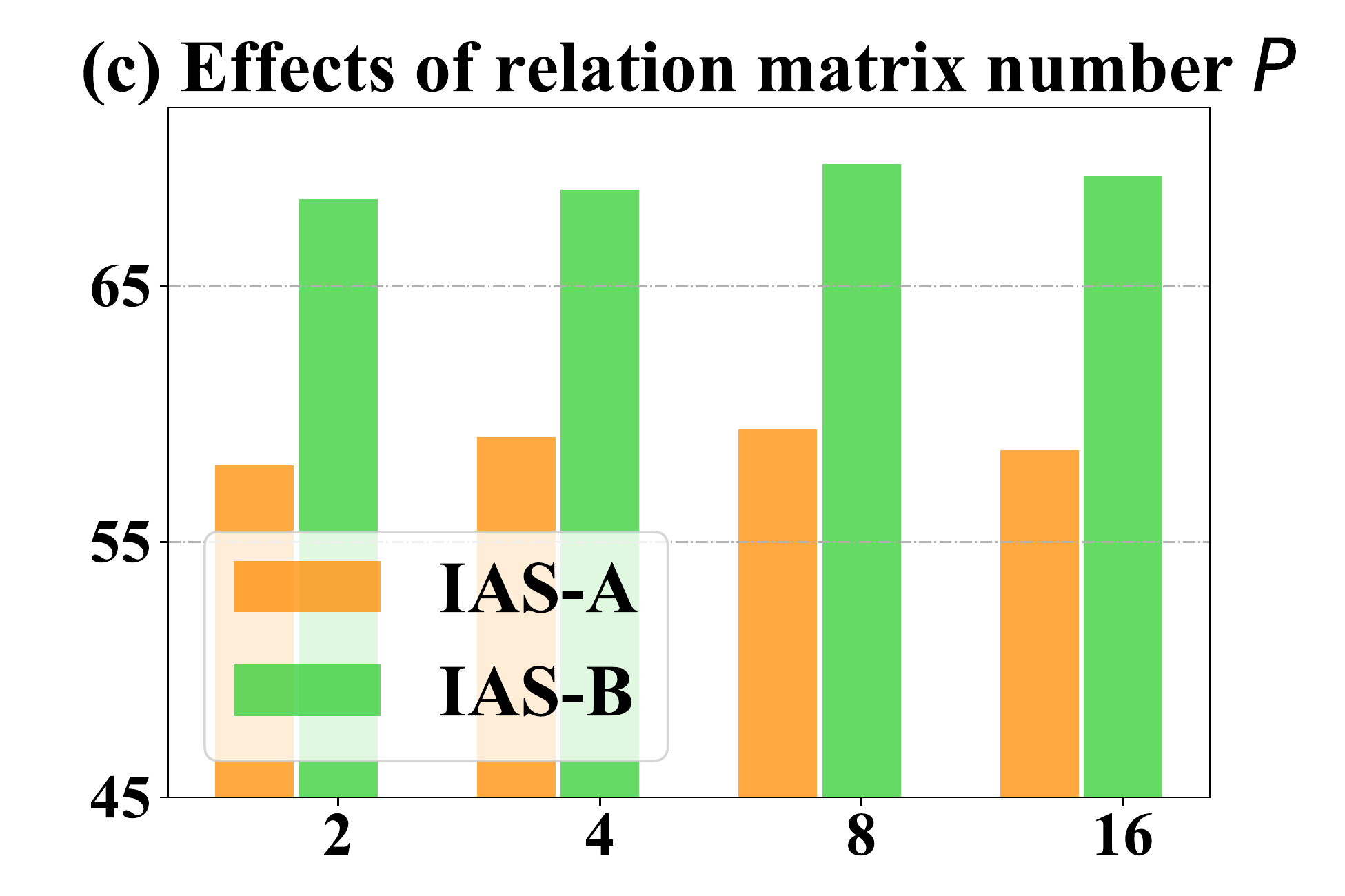}}\ \ \ 
       \scalebox{0.15}{\includegraphics[]{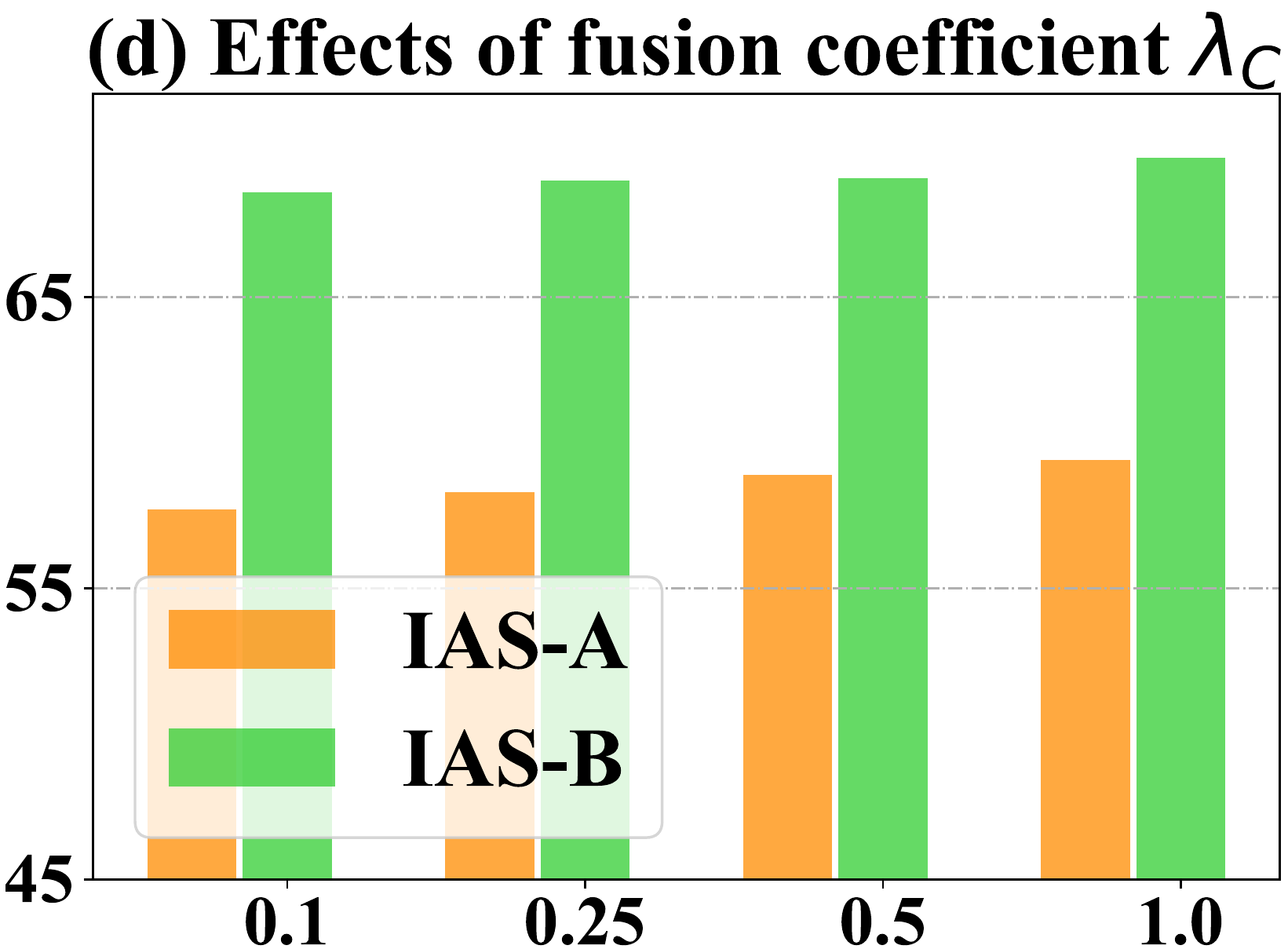}} \ \ \
         \scalebox{0.15}{\includegraphics[]{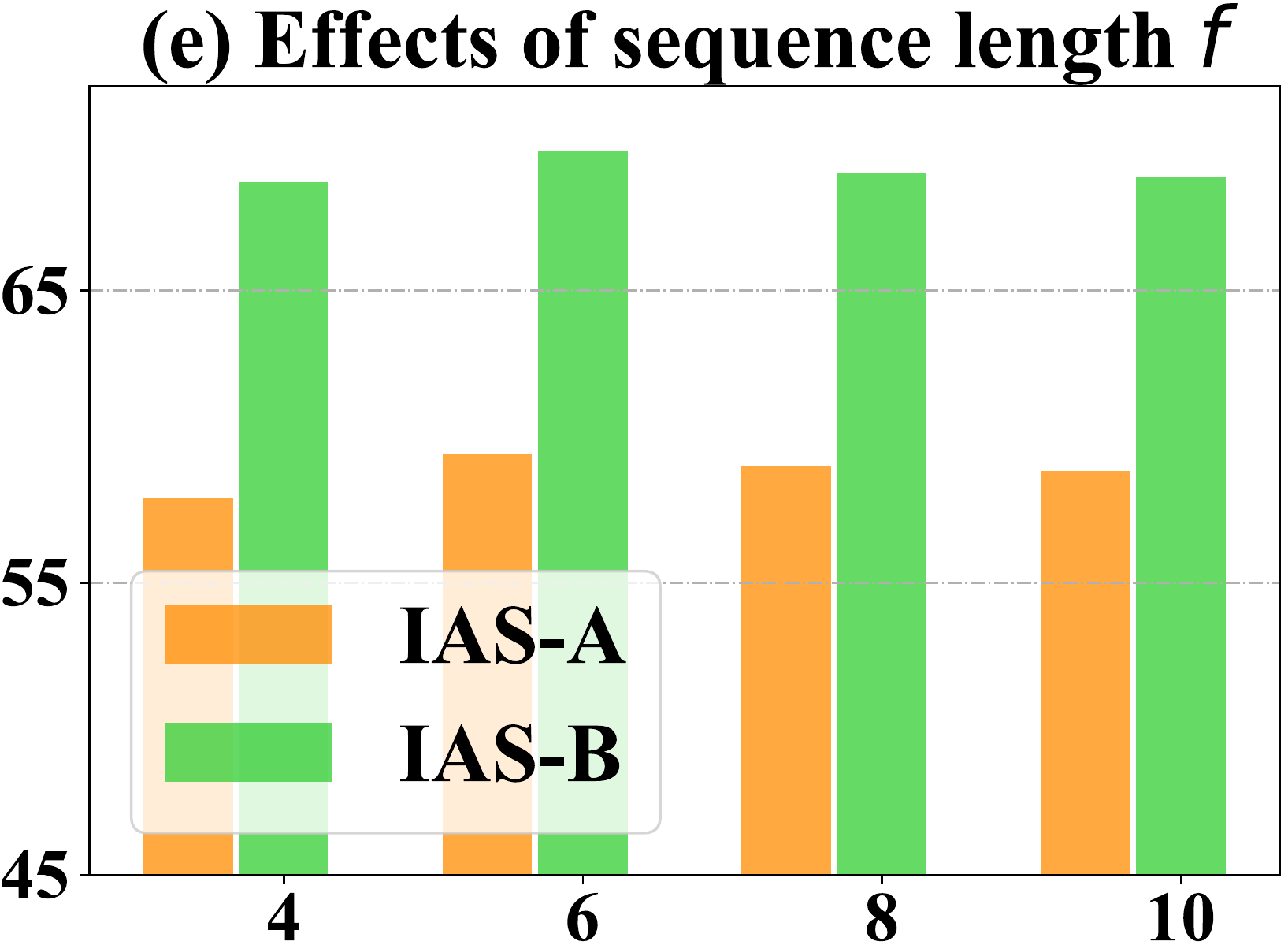}}
         \ \ \
         \scalebox{0.15}{\includegraphics[]{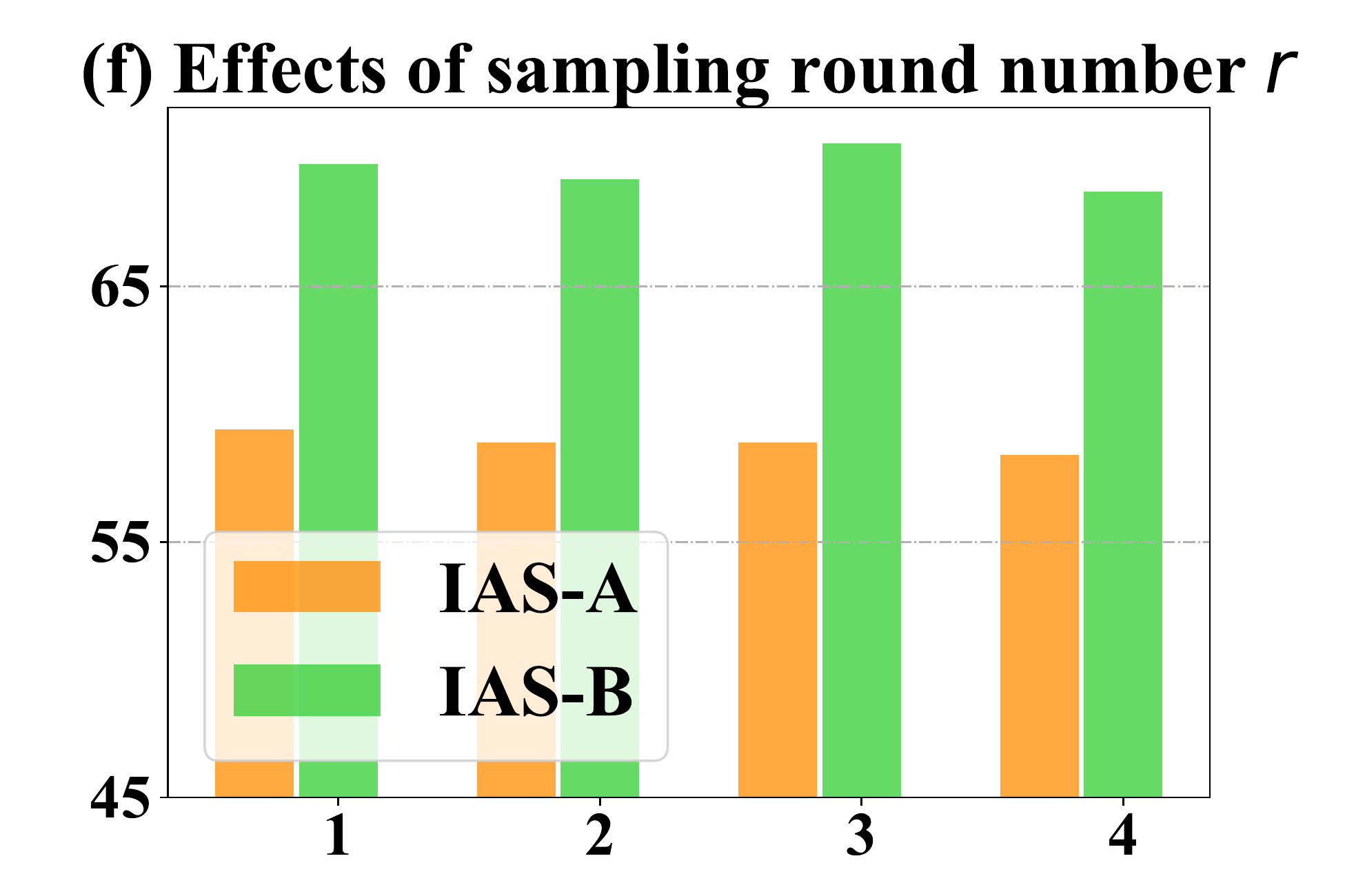}}
    \caption{Rank-1 accuracy on IAS-Lab illustrating effects of different hyper-parameters: (a)-(b) Temperature $T_1$ and $T_2$. (c) Structural relation matrix number $P$. (d) Collaboration fusion coefficient $\lambda_{C}$. (e) Sequence length $f$. (e) Random sampling round number $r$. Zoom in for the better visualization.}
    \label{parameters}
\end{figure}

\begin{table}[t]
\caption{Performance of SM-SGE with different graph
scales. }
\label{graph_scales}
\scalebox{0.75}{
\setlength{\tabcolsep}{2.6mm}{
\begin{tabular}{cccccc}
\specialrule{0.1em}{0.45pt}{0.45pt}
\textbf{Body-scale} & \textbf{Part-scale} & \textbf{Joint-scale} & \textbf{\begin{tabular}[c]{@{}c@{}}Hyper-joint\\ -scale\end{tabular}} & \textbf{IAS-A} & \textbf{IAS-B} \\ \specialrule{0.1em}{0.45pt}{0.45pt}
\checkmark                  &                     &                      &                                                                       & 53.3           & 60.4           \\
\checkmark                   & \checkmark                   &                      &                                                                       & 56.9           & 65.8           \\
\checkmark                   & \checkmark                   & \checkmark                    &                                                                       & 58.3           & 68.5           \\
\checkmark                   & \checkmark                   & \checkmark                    & \checkmark                                                                     & \textbf{59.4}  & \textbf{69.8}  \\ \specialrule{0.1em}{0.2pt}{0.2pt}
\end{tabular}
}
}
\end{table}

\textbf{Effects of Multiple Graph Scales.} 
As shown in Table \ref{graph_scales}, combining graph scales from coarse (body-scale) to fine (hyper-joint-scale) progressively improves Rank-1 accuracy by $6.1\%$-$9.4\%$ on both IAS-A and IAS-B. Compared with the plain reconstruction of body-scale graphs (note that plain reconstruction without collaborative relation learning is employed when using only body-scale graphs, shown in first row), employing multi-scale skeleton reconstruction (MSR) based on two adjacent scales of graphs (body-scale and part-scale) obtains a significant performance gain by $3.6\%$-$5.4\%$ Rank-1 accuracy. These results further demonstrate the effectiveness of multi-scale graphs and MSR, which are able to capture more discriminative skeleton features at 
various levels for person Re-ID.

\textbf{Model Sensitivity Analysis.} We evaluate effects of different hyper-parameters ($T_1,T_2,P,\lambda_{C},f, r$) on SM-SGE: \textbf{(1)}
We observe that SM-SGE is not sensitive to temperature changes from 0.1 to 1.0 (see Fig. \ref{parameters} (a), (b)). Since lower temperatures tend to ignore more similar information \cite{hinton2015distilling} and could reduce relation learning performance, we set temperature to 1.0 for all relation learning in our framework.
\textbf{(2)} As shown in Fig. \ref{parameters} (c), introducing more learnable structural relation matrices with larger $P$ can improve model performance on both IAS-A and IAS-B. However, too many relation matrices ($P=16$) may cause the model to learn redundant relation information, which leads to a slight performance degradation. \textbf{(3)}
The parameter $\lambda_{C}$ controls the degrees of fusion with collaborative node features. We find that fusing graph node features with larger $\lambda_{C}$ can achieve better Re-ID performance (see Fig. \ref{parameters} (d)), which verifies the necessity of sufficient multi-scale collaboration fusion to learn a more effective skeleton representation. \textbf{(4)} As shown in Fig. \ref{parameters} (e) (f), SM-SGE obtains the highest Re-ID performance when $f=6$ and $r=1$ in most cases. Although larger $f$ and $r$ can provide more training subsequences ($i.e.,$ up to $r(f-1)$ samples), it can increase the computation complexity ($e.g.,$ requires more computation memory and time for training), thus we set $f=6$ and $r=1$ to achieve better trade-off between performance and complexity.


 \begin{figure}[t]
    \centering
    \quad  \subfigure[\scalebox{0.9}{CCR Visualization (KS20)}]{    \scalebox{0.265}{\label{KS20_CCR}\includegraphics[]{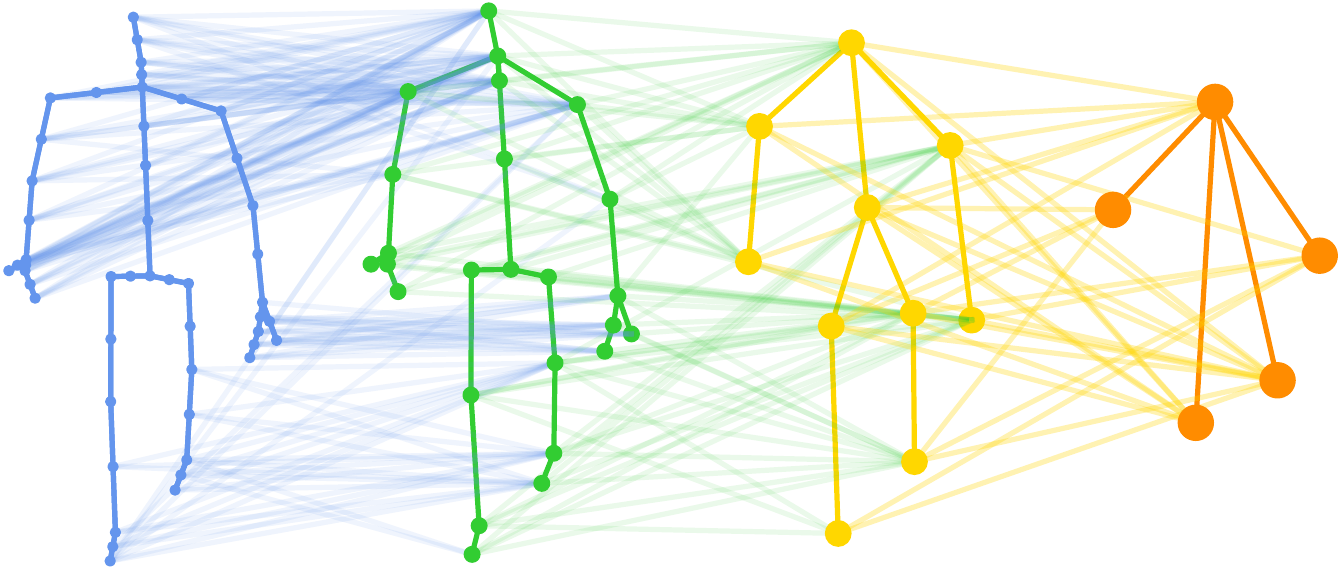}}
     }
     \quad \subfigure[\scalebox{0.9}{CCR Visualization (IAS-A)}]{    \scalebox{0.235}{\label{IAS_CCR}\includegraphics[]{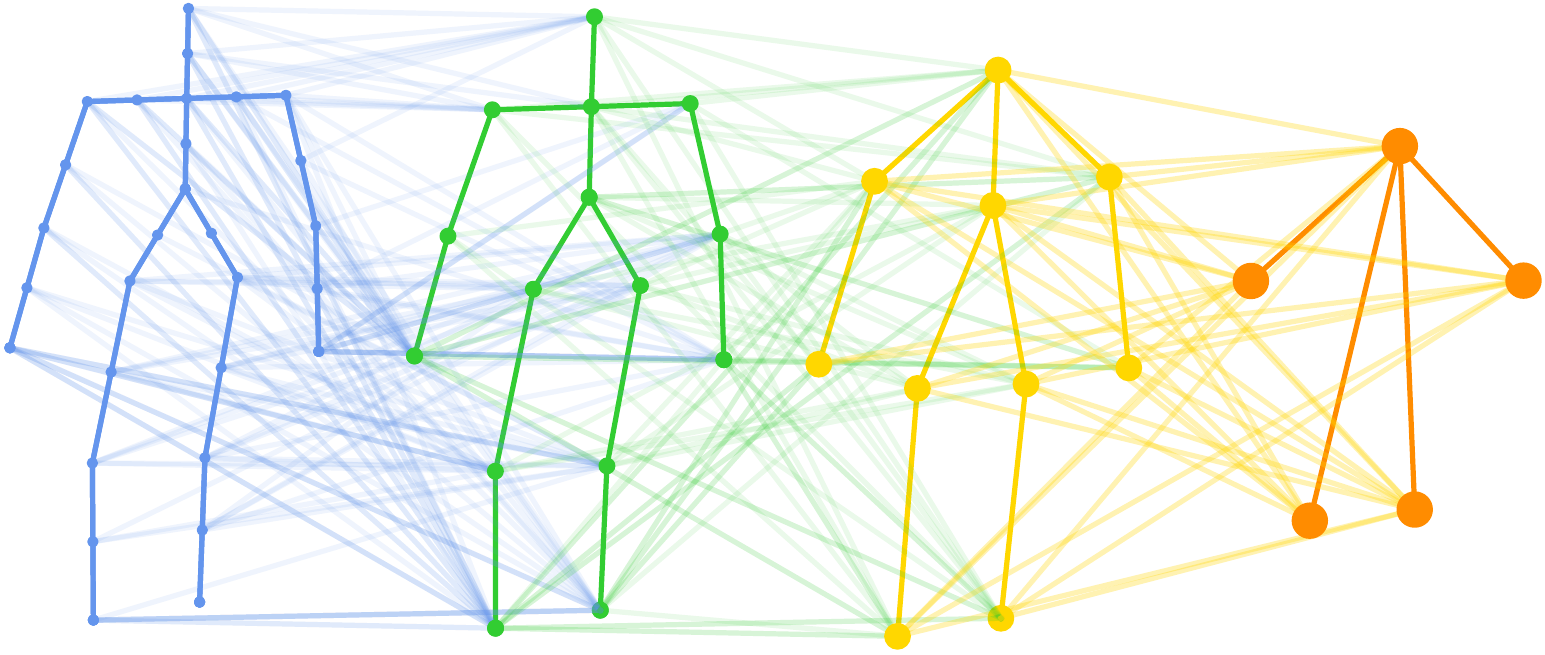}}
     }
    \ \
    \subfigure[\scalebox{0.9}{CCR Matrices (KS20)}]{\scalebox{0.1}{\label{KS20_matrix}\includegraphics[]{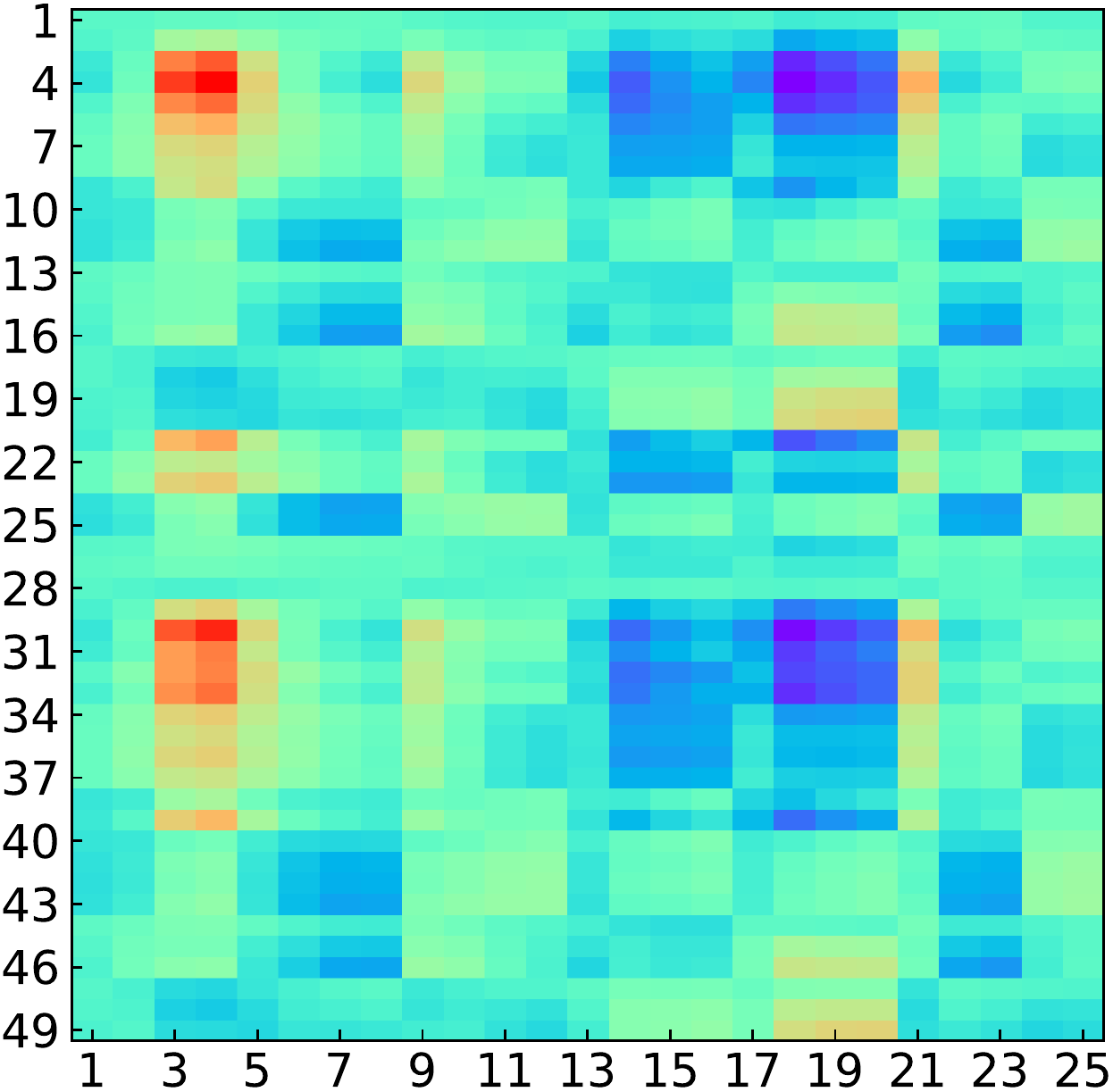}} \scalebox{0.1}{\includegraphics[]{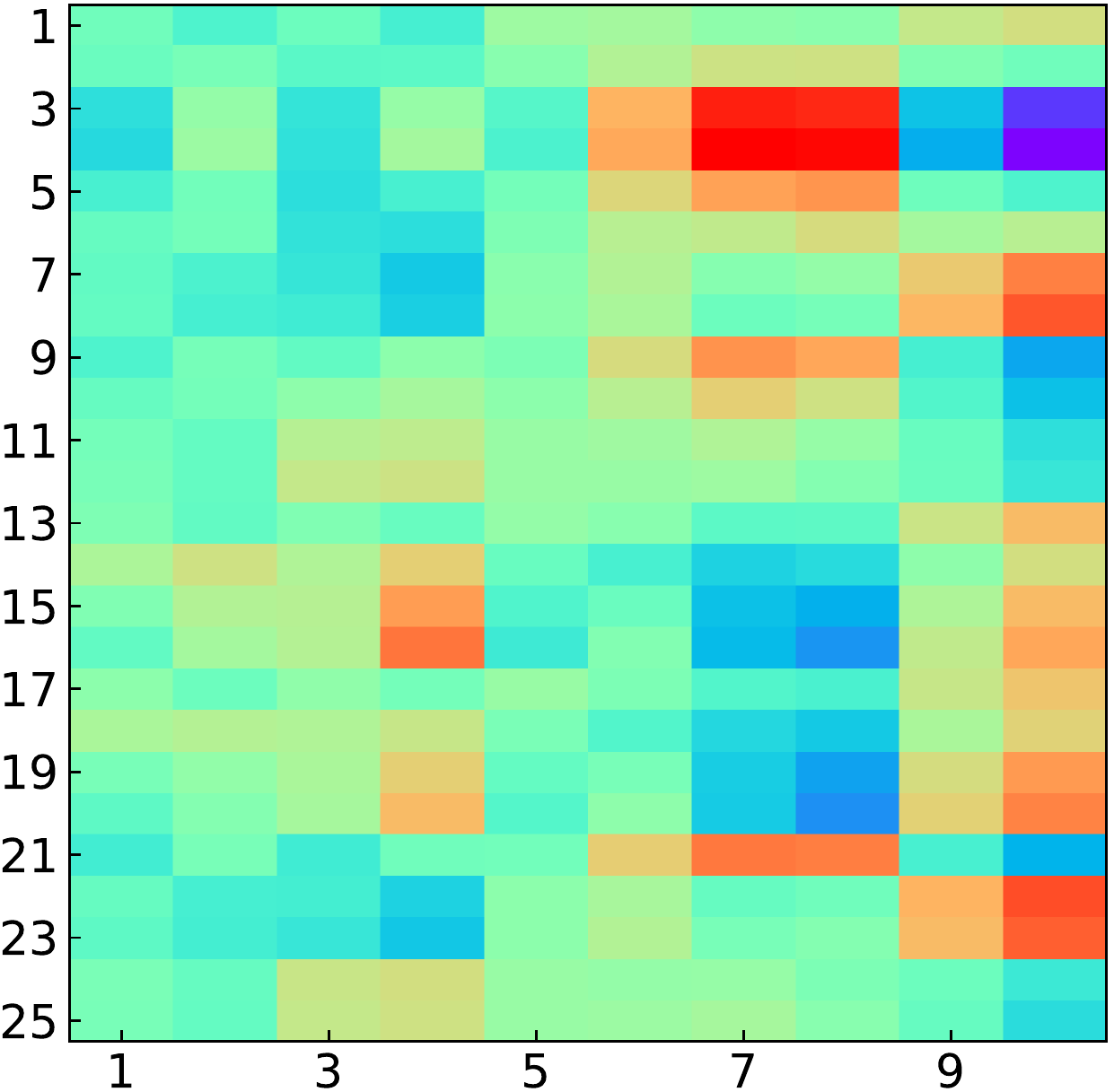}}
    \scalebox{0.1}{\includegraphics[]{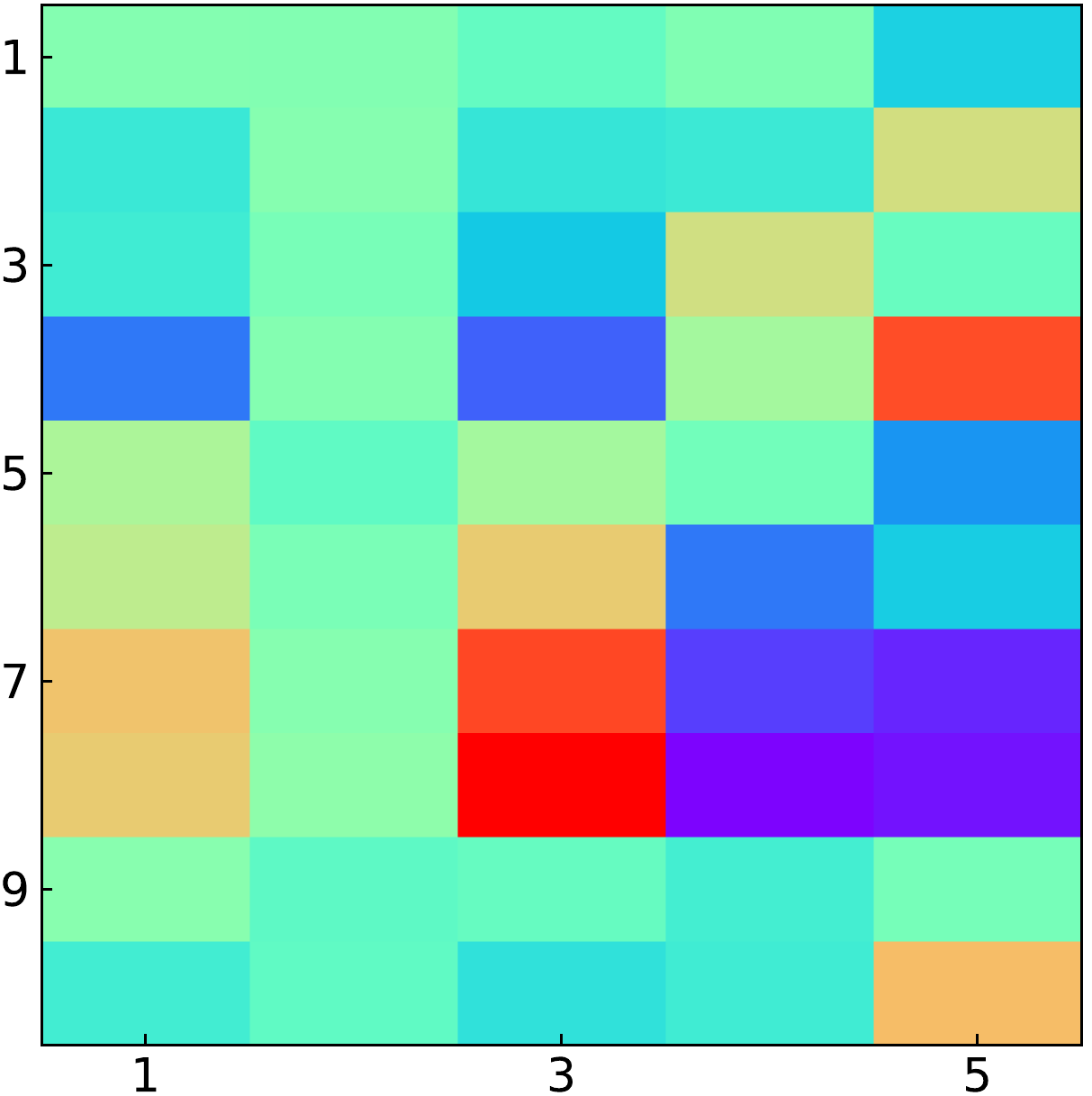}}
    }
    \ \ 
    \subfigure[\scalebox{0.9}{CCR Matrices (IAS-A)}]{\scalebox{0.1}{\label{IAS_matrix}\includegraphics[]{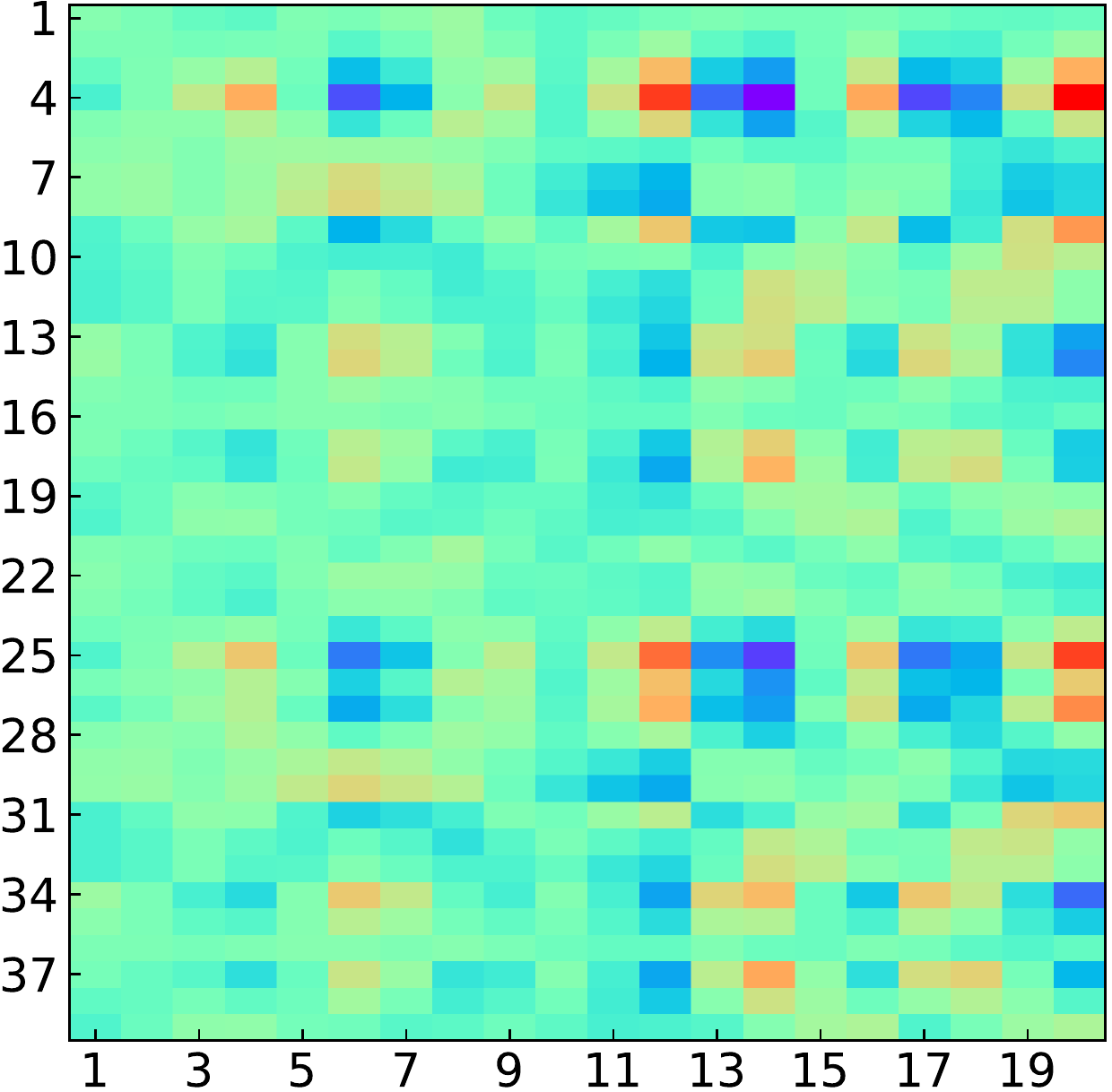}} \scalebox{0.1}{\includegraphics[]{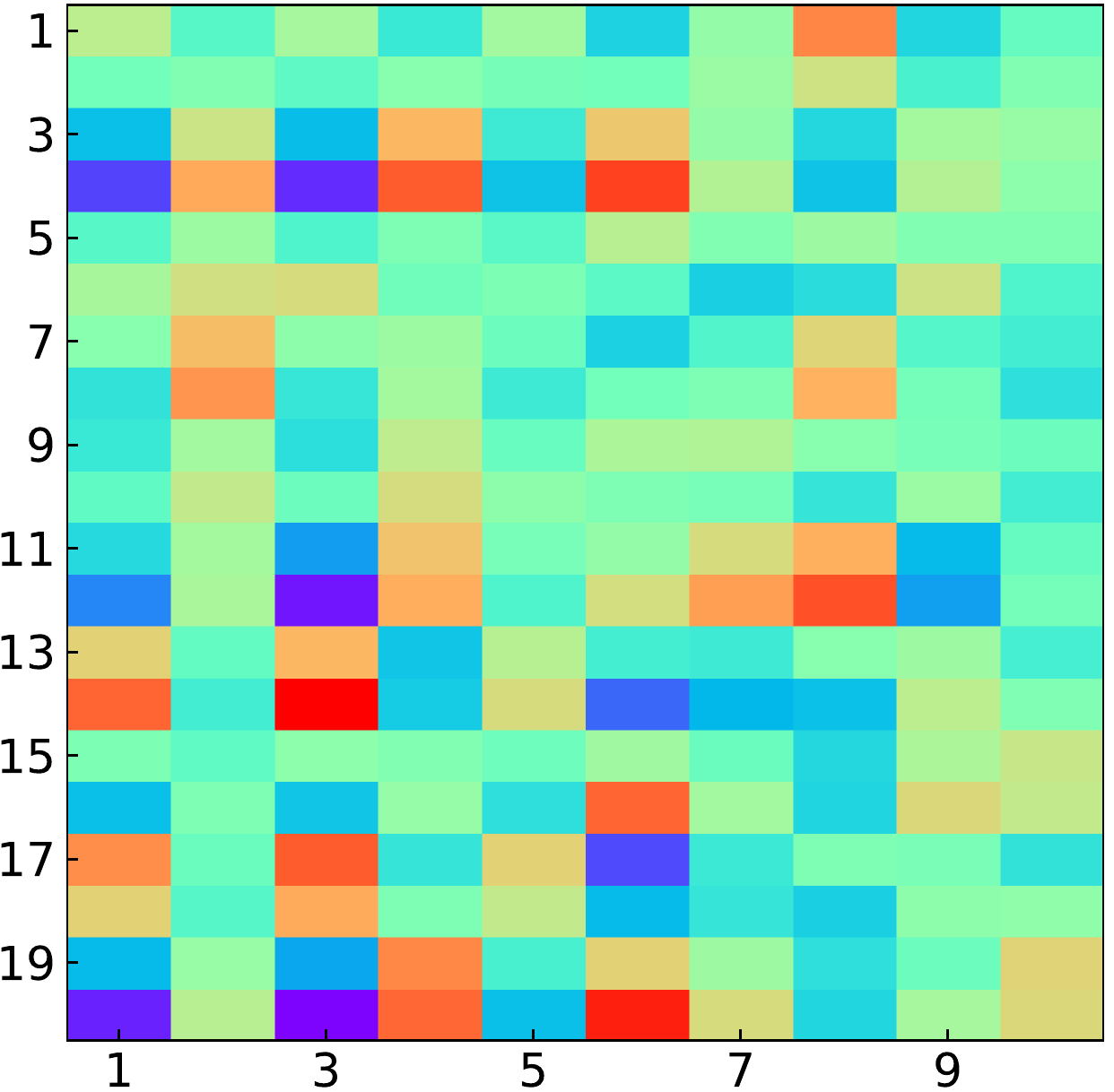}}
    \scalebox{0.1}{\includegraphics[]{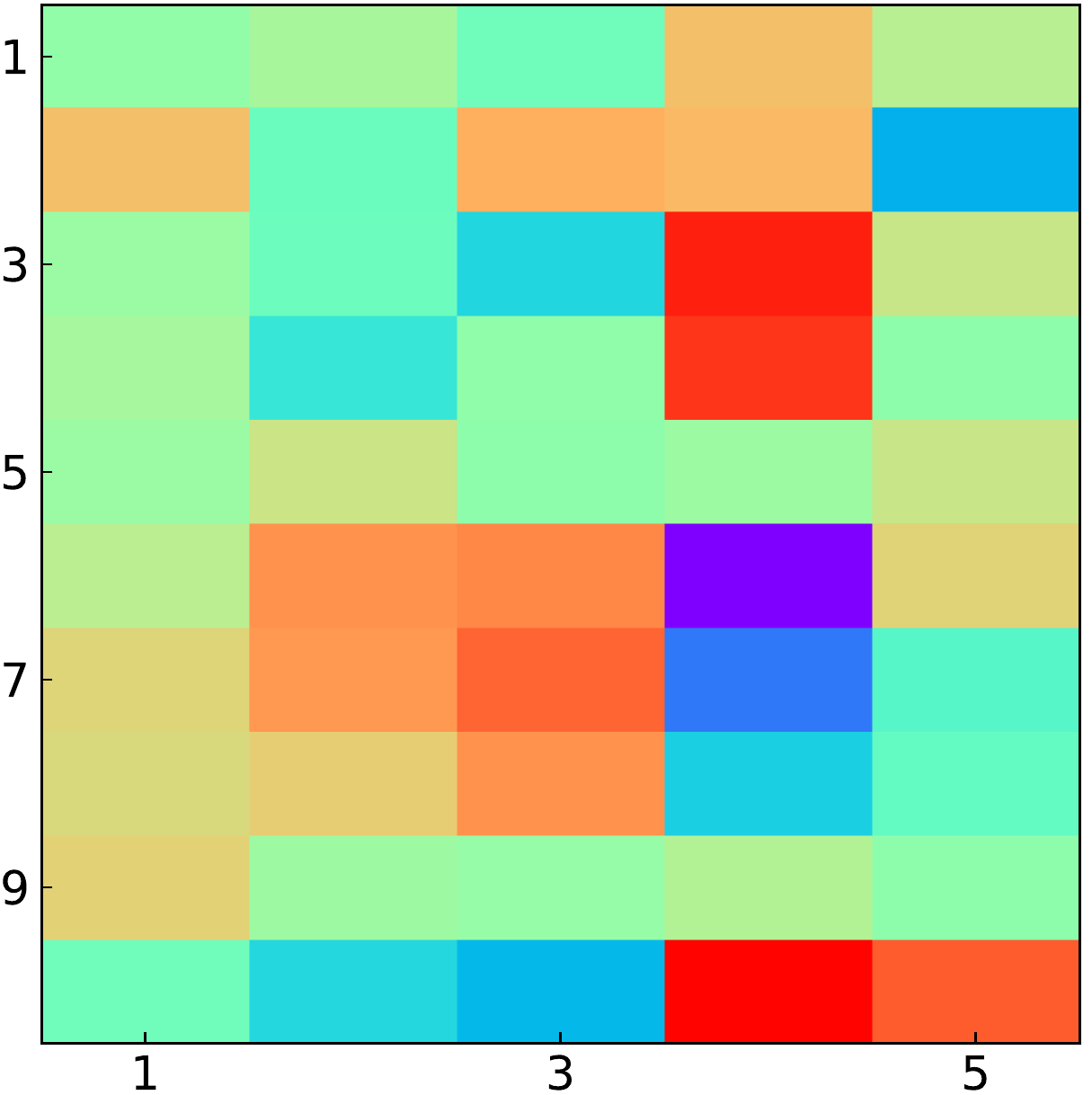}}
    }
    \caption{(a)-(b): Cross-scale collaborative relations (CCR) among body components for sample skeletons in KS20 and IAS-A. (c)-(d): CCR matrices ($\widehat{\mathbf{A}}^{0, 1},\widehat{\mathbf{A}}^{1, 2},\widehat{\mathbf{A}}^{2, 3}$) for (a) and (b). Note that the abscissa and ordinate denote indices of nodes. }
    \label{visual}
\end{figure}

\textbf{Analysis of Cross-scale Collaborative Relations.} 
We visualize positions of different body components and their collaborative relations across adjacent scales (note that we draw significant relations with values larger than $90\%$ maximum value of $\widehat{\mathbf{A}}^{m-1, m}$), and we obtain observations as follows: \textbf{(1)} As shown in Fig. \ref{KS20_CCR} and \ref{IAS_CCR}, spatially corresponding or nearby body components ($e.g.,$ subcomponents of limbs at the same side) possess evident relations across different scales, which demonstrates that SM-SGE can capture the high-level semantics of body-component correspondence between different graphs. \textbf{(2)} For non-adjacent body components ($e.g.,$ arms, legs) with a joint movement trend, the framework also learns higher correlations among their corresponding nodes in different graphs (see Fig. \ref{KS20_matrix}, \ref{IAS_matrix}), which justifies our claim that SM-SGE framework is able to adaptively infer global body-component cooperation in skeletal motion. More results and proofs are provided in Appendix.

\section{Conclusion}
In this paper, we model 3D skeletons as multi-scale graphs, and propose a self-supervised multi-scale skeleton graph encoding (SM-SGE) framework to learn an effective representation from unlabeled skeleton graphs for person Re-ID. To capture key correlative features of graph nodes, we propose the multi-scale graph relation network (MGRN) to learn structural and collaborative relations among body-component nodes in different graphs.
A novel multi-scale skeleton reconstruction (MSR) mechanism with subsequence reconstruction and cross-scale skeleton inference tasks is devised to encode graph dynamics and discriminative high-level features of skeleton graphs for person Re-ID. SM-SGE outperforms most state-of-the-art skeleton-based methods, and it can achieve satisfactory performance on 3D skeleton data estimated from RGB videos.

\bibliographystyle{ACM-Reference-Format}
\bibliography{acmart.bib}
\end{document}